\documentclass{article}

\usepackage[margin=1.0in]{geometry}
\usepackage[utf8]{inputenc}
\usepackage{graphicx}
\usepackage[numbers]{natbib}

\usepackage[utf8]{inputenc} 
\usepackage[T1]{fontenc}    
\usepackage{lmodern}
\usepackage{hyperref}       
\usepackage{url}            
\usepackage{booktabs}       
\usepackage{amsfonts}       
\usepackage{nicefrac}       
\usepackage{microtype}      
\usepackage{amsmath}
\usepackage{caption}
\usepackage{subcaption}

\usepackage{xcolor}

\newcommand{\es}[2] {\begin{align} \label{#1} \begin{split} #2 \end{split} \end{align}}

\newtheorem{lemma}{Lemma}

\graphicspath{ {./Figures/Main/} }

\title{\textbf{Anatomy of Catastrophic Forgetting: Hidden Representations and Task Semantics}}
\date{}

\author{
Vinay Ramasesh\\
\texttt{ramasesh@google.com}\\
Google\\
Mountain View, CA
\and
Ethan Dyer\\
\texttt{edyer@google.com}\\
Google\\
Mountain View, CA
\and
Maithra Raghu\\
\texttt{maithrar@gmail.com}\\
Google\\
Mountain View, CA
}

\begin{document}

\maketitle

\begin{abstract}
A central challenge in developing versatile machine learning systems is \textit{catastrophic forgetting} --- a model trained on tasks in sequence will suffer significant performance drops on earlier tasks.
Despite the ubiquity of catastrophic forgetting, there is limited understanding of the underlying process and its causes. In this paper, we address this important knowledge gap, investigating how forgetting affects representations in neural network models. Through representational analysis techniques, we find that \textit{deeper layers} are disproportionately the source of forgetting. Supporting this, a study of methods to mitigate forgetting illustrates that they act to stabilize deeper layers. These insights enable the development of an analytic argument and empirical picture relating the degree of forgetting to representational similarity between tasks. Consistent with this picture, we observe maximal forgetting occurs for task sequences with \textit{intermediate} similarity. We perform empirical studies on the standard split CIFAR-10 setup and also introduce a novel CIFAR-100 based task approximating realistic input distribution shift.
\end{abstract}
\vspace{.2in}
\section{Introduction}
While the past few years have seen the development of increasingly versatile machine learning systems capable of learning complex tasks \cite{stokes2020deep, raghu2020survey, kostrikov2020image, raffel2019exploring, wu2019detectron2}, \textit{catastrophic forgetting} remains a core challenge to designing machine learning systems. At a high level, catastrophic forgetting is the ubiquitous phenomena that machine learning models trained on non-stationary data distributions suffer performance losses on older data instances. More specifically, if our machine learning model is trained on a sequence of tasks, accuracy on earlier tasks drops significantly.
The catastrophic forgetting problem manifests in many sub-domains of machine learning including continual learning \cite{kirkpatrick2017overcoming}, multi-task learning and transfer \cite{Kudugunta2019InvestigatingMN} and even in standard supervised learning through input distribution shift \cite{Toneva2019AnES, snoek2019can, rabanser2019failing, recht2019imagenet, arivazhagan2019massively} and data augmentation \cite{gontijo2020affinity}.

Catastrophic forgetting has been an important focus of recent research in the context of deep learning, with numerous methods proposed to help mitigate forgetting \cite{goodfellow2013empirical, kirkpatrick2017overcoming, lee2017overcoming, li2019learn, serra2018overcoming, ritter2018online, rolnick2019experience}. However, many of these methods are only effective in specific settings \cite{kemker2018measuring}, and fully mitigating forgetting across the diverse scenarios it arises in is hindered by a lack of understanding of the fundamental properties of catastrophic forgetting.

How does catastrophic forgetting affect the hidden representations of neural networks? Are earlier tasks forgotten equally across all parameters? Are there underlying principles common across methods to mitigate forgetting? How is catastrophic forgetting affected by (semantic) similarities between sequential tasks? And what are good benchmark tasks that capture the essence of how catastrophic forgetting naturally arises in practice? This paper begins to provide answers to these and other questions. Our contributions are as follows:
\begin{itemize}
\item We perform a thorough empirical investigation of how catastrophic forgetting affects the hidden representations of neural networks, by using representational similarity measures \cite{kornblith2019similarity, raghu2017svcca}, layer freezing and layer reset experiments \cite{zhang2019all}. We find that deeper layers are the primary source of forgetting --- changing the most due to sequential training.

\item These studies and all subsequent experiments are performed on a standard split CIFAR-10 task, and a novel CIFAR-100 based task we introduce, that approximates input distribution shift (e.g. in online learning) by using the hierarchical label structure of the CIFAR-100 images.

\item We then examine the effect on layer representations of two popular but different approaches for mitigating forgetting: replay buffers \cite{Ratcliff1990ConnectionistMO, rolnick2019experience} and elastic weight consolidation \cite{kirkpatrick2017overcoming}, finding that these methods mitigate forgetting by stabilizing deeper representations. 

\item We provide empirical evidence that forgetting follows semantically consistent patterns, with the degree of forgetting related to task similarity. But crucially, we find accurate measures of task similarity (which are connected to network representations) depend on both the data \textit{and} the optimization process.

\item We formalize the effect of semantic similarity on forgetting through a simple analytic framework, which, consistent with experiments, reveals that forgetting is most severe for tasks with intermediate similarity.

\end{itemize}
\section{Related Work}
Much of the research on catastrophic forgetting in deep learning has concentrated on developing methods for mitigating forgetting, with techniques to implicitly or explicitly partition parameters for different tasks \cite{goodfellow2013empirical, cheung2019superposition, rusu2016progressive, golkar2019continual}, regularization (elasticity) and Bayesian approaches \cite{kirkpatrick2017overcoming, lee2017overcoming, zenke2017continual, li2017learning, ritter2018online, ebrahimi2019uncertainty} as well as directly storing data from earlier tasks in replay buffers \cite{rolnick2019experience, shin2017continual, isele2018selective, Ratcliff1990ConnectionistMO}. However, there has been limited analysis on how forgetting is caused by different hidden representations in the neural network, which is a key focus of this paper. 

Other prior work on understanding catastrophic forgetting has looked at properties of task sequences \cite{nguyen2019toward} or as a means of comparing different neural network architectures \cite{arora2019does}. Many methods also use benchmarks like permuted-MNIST, which, while providing a useful task for initial study, may lead to mitigation methods working only in limited settings \cite{kemker2018measuring}. To gain more generalizable insights, we focus on a split CIFAR-10 task, and also introduce a novel CIFAR-100 based task, which approximates realistic input distribution shift --- a common way forgetting arises in practice \cite{recht2019imagenet, arivazhagan2019massively}.

\section{Datasets and Tasks for Catastrophic Forgetting}
\label{sec-datasets-tasks}
To study the key characteristics of catastrophic forgetting and identify generalizable insights, we must define realistic tasks that capture ways in which catastrophic forgetting appears in practice. To this end, we perform experiments on both the standard split CIFAR-10 task and a novel CIFAR-100 based task which approximates input distribution shift, across multiple neural network architectures.

\textbf{Split CIFAR-10 Task:} This task consists of splitting the standard ten class image classification of CIFAR-10 into $m$ sequential tasks of $n$ classes each. This work uses $m=2$ tasks, each task a disjoint set of five classes. The underlying neural network model usually consists of $m$ classification heads, one for each task, along with a shared \textit{body}. This is a standard setup for studying catastrophic forgetting, and has also been used with the smaller-scale permuted/split MNIST tasks ~\cite{zenke2017continual, kirkpatrick2017overcoming}. 

\textbf{Split CIFAR-100 Distribution Shift Task:} While the split CIFAR-10 task with its task specific neural network heads is a standard benchmark for catastrophic forgetting, it doesn't directly represent the scenario where catastrophic forgetting arises through \textit{input distribution shift} --- the input data to the neural network undergoes a distribution shift, causing the neural network to perform poorly on earlier data distributions \cite{arivazhagan2019massively, snoek2019can, rabanser2019failing}. To certify that our conclusions also generalize to this highly important case, we introduce a CIFAR-100 based task to approximate input distribution shift. Here an individual task consists of distinguishing between $n$ CIFAR-100 \textit{superclasses}, which have training/test (input) data corresponding to a \textit{subset} of their constituent classes. Different tasks have different subsets of constituent classes making up the input data, giving an input distribution shift across tasks. For a specific example, see Figure~\ref{fig:semanticscifar100} and Appendix \ref{sec-SuppCIFAR-100}.

\textbf{Neural Network Architectures:} In addition to these two different tasks (representing distinct ways in which catastrophic forgetting might arise), we perform experiments across a range of neural network architectures --- VGG~\cite{simonyan2014deep}, ResNet~\cite{he2015deep} and DenseNet~\cite{huang2016densely}. The presence of (increasing numbers of) \textit{skip-connections} in the latter two is an important test of how generalizable our forgetting insights are across architectures.

\begin{figure}
\centering
\hspace*{-8mm}\includegraphics[width=14.72cm]{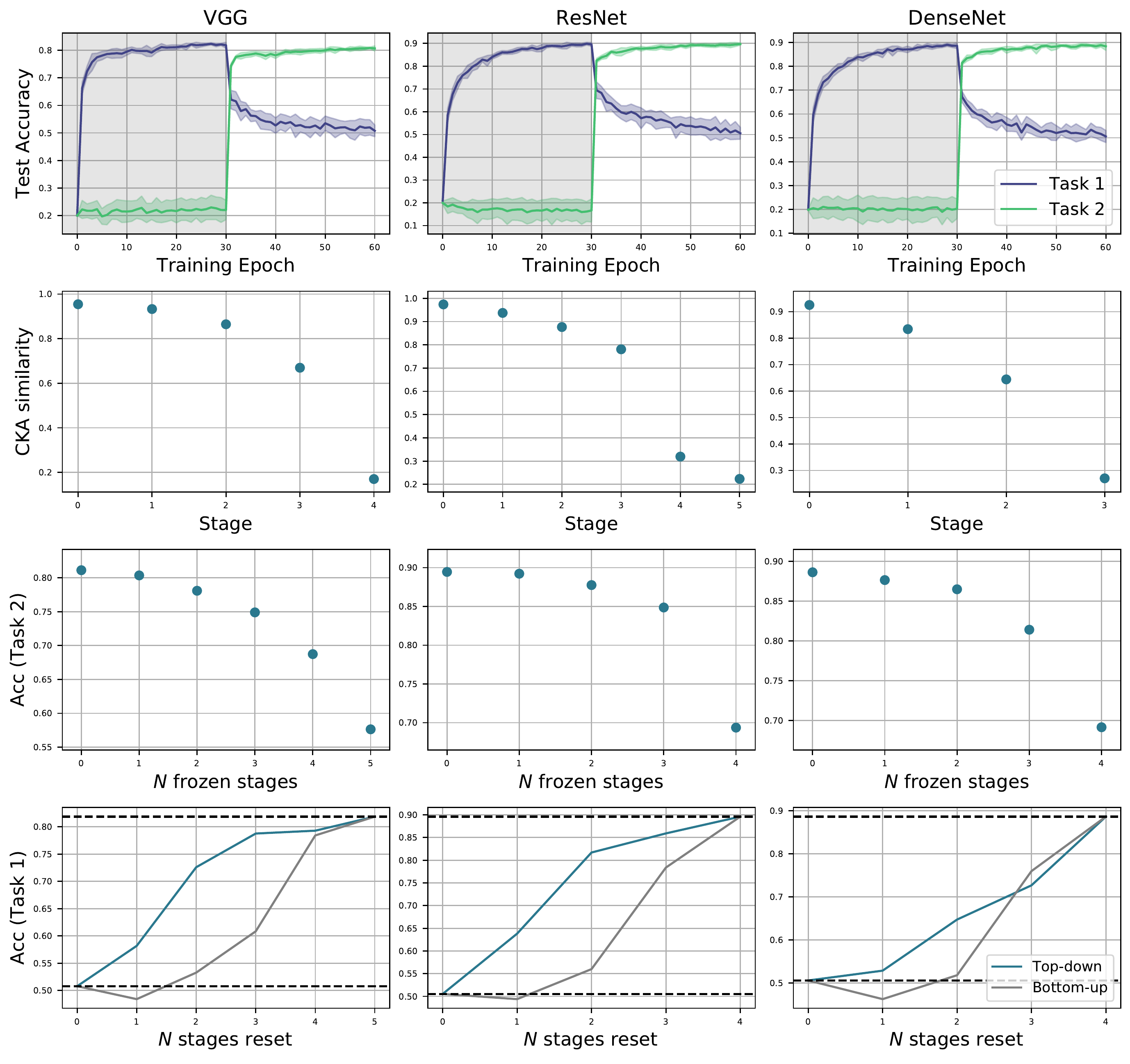}
\caption{\small \textbf{Higher layers are the primary source of catastrophic forgetting on split CIFAR-10 task.}
Test accuracy for Task 1 (five classes of CIFAR-10) and Task 2 (remaining five classes of CIFAR-10) during sequential training (top row).
Representational similarity (CKA) scores between activations of stages (blocks of convolutions) before and after training on Task 2 indicate that lower layers don't change much through training on task two, while higher layers change significantly (second row).
When freezing a contiguous block of layers (starting from lowest) and measuring accuracy during training on Task 2, we observe lower layers can be frozen with little impact (third row).
Finally, after training on Task 2 we reset contiguous blocks of layers to their values before training and record the resulting accuracy on Task 1 (bottom row). We see a significant increase in accuracy when resetting the highest $N$ layers (blue line) compared to resetting the $N$ lowest layers (gray line). Together, these results demonstrate that higher layers are disproportionately responsible for forgetting. Further experimental details can be found in Appendix~\ref{sec:app_methods} with additional experiments in Appendix~\ref{sec:app_resetretrain}.
}
\label{fig-hiddenreps-cifar10}
\end{figure}

\section{Catastrophic Forgetting and Neural Network Hidden Representations}
\label{sec-forgetting-hidden-reps}
Having defined the datasets and tasks for studying catastrophic forgetting, we turn to understanding its effects on neural network hidden representations. Specifically, we investigate whether representations throughout the network are equally `responsible' for forgetting, or whether specific layers forget more than others through the sequential training process.

Letting \textit{higher layers} refer to layers closest to the output, and \textit{lower layers} refer to layers closer to the input, our empirical results clearly demonstrate that higher layers are the primary source of forgetting. By contrast, many of the lower layers remain stable through the sequential training process. These results are evident in both split CIFAR-10 and the CIFAR-100 distribution shift task. To measure the degree of forgetting across different layers, we turn to (i) representational similarity techniques (ii) layer freezing experiments, and (iii) layer reset experiments \cite{zhang2019all}. 

\begin{figure}
\centering
\hspace*{-8mm} \includegraphics[width=14.72cm]{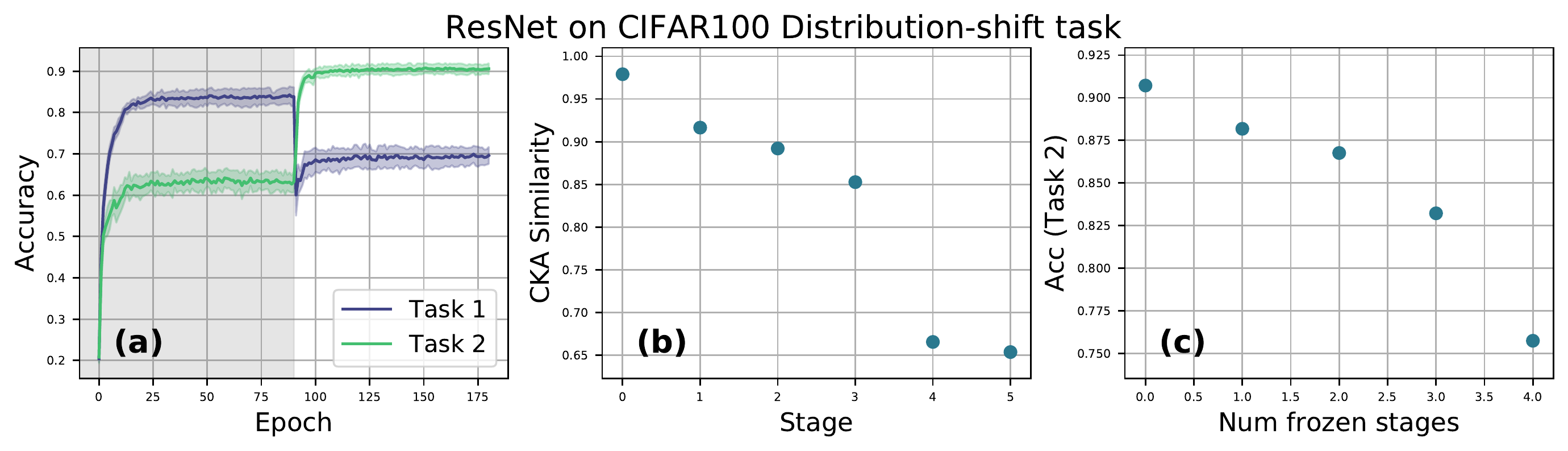}
\caption{ \small \textbf{Higher layers also drive catastrophic forgetting on the distribution shift CIFAR-100 task.} The figure shows (a) training curves, (b) CKA representational similarity and (c) layer freezing experiments on the ResNet model trained on the CIFAR-100 distribution shift task, with additional consistent VGG and DenseNet architectures results in Appendix~\ref{sec:app_CIFAR-100exprmts}. As in Figure \ref{fig-hiddenreps-cifar10}, we use CKA to measure the similarity between hidden layer representations before and after training on Task 2. Again, we find lower layers change less than higher layers through training (b). In pane (c) we freeze a contiguous block of layers (starting from the lowest layers) when training on Task 2 and record accuracy on Task 2. We find that freezing the lowest layers does not significantly affect Task 2 accuracy.
}
\label{fig-hiddenreps-cifar100}
\end{figure}

Representational similarity techniques take in a pair of neural network (hidden) layer representations and output a similarity score between $0$ and $1$ depicting how similar the representations are \cite{kornblith2019similarity, raghu2017svcca, morcos2018insights}. In our setting, we compare layer representations before and after training (sequentially) on the second task, and we measure similarity using centered kernel alignment (CKA) \cite{kornblith2019similarity}. See Appendix~\ref{sec:app_methods} for full details of our implementation. The results across the two tasks and multiple network architectures are shown in Figures \ref{fig-hiddenreps-cifar10}, \ref{fig-hiddenreps-cifar100}, where we observe that lower layers representations remain stable through training on the second task, whereas higher layer representations change significantly.

Further evidence that the change in higher layer representations is the source of forgetting is given through layer freezing and layer reset experiments. In the layer freezing experiments, we freeze the parameters of a contiguous block of layers, starting from the lowest layers, when training on the second task. We observe that freezing the lowest few layers has little effect on accuracy on the second task, illustrating that the lowest layer representations are not forgetting (changing) during the sequential training process. 

In layer resets, we pinpoint this effect even further. Having trained on the second task, we \textit{reset} a contiguous block of layers to their values at the end of training on the first task. Doing this for blocks of layers starting from the highest layer (excluding the task output head) gives in a significant increase in accuracy in the first task. This further demonstrates that forgetting is driven by the highest layers.  Appendices~\ref{sec:app_linreg} and \ref{sec:app_resetretrain} present two more types of experiments---(i) linear classification on forgotten activations and (ii) retraining with frozen layers---supporting this conclusion.

\section{Forgetting Mitigation Methods and Representations}
In this section we turn to investigating the key properties of methods to mitigate forgetting. The results of Section \ref{sec-forgetting-hidden-reps} might suggest that the only approach is to stabilize higher layer representations. But surprisingly, alternate mitigation methods such as weight orthogonalization can tolerate arbitrarily large changes in higher layers --- we present an example in Appendix~\ref{sec:app_analytic}. Here we ask whether common methods to mitigate forgetting stabilize higher layers or evade forgetting in a different way.

There are many diverse methods that have been proposed to mitigate forgetting, from which we choose two (i) elastic weight consolidation (EWC)~\cite{kirkpatrick2017overcoming}, which nudges parameters to remain close to their previously-trained values by imposing a quadratic penalty on deviations (details in Appendix~\ref{SuppEWC}) and (ii) replay buffers ~\cite{Ratcliff1990ConnectionistMO}, which retains a subset (buffer) of old task training data, which it periodically reuses (replays) during training on subsequent tasks. These two methods and their variants are popular, successful and importantly, represent two very different approaches to mitigate forgetting.

\begin{figure}
\centering
\hspace*{-5mm}
\includegraphics[width=14cm]{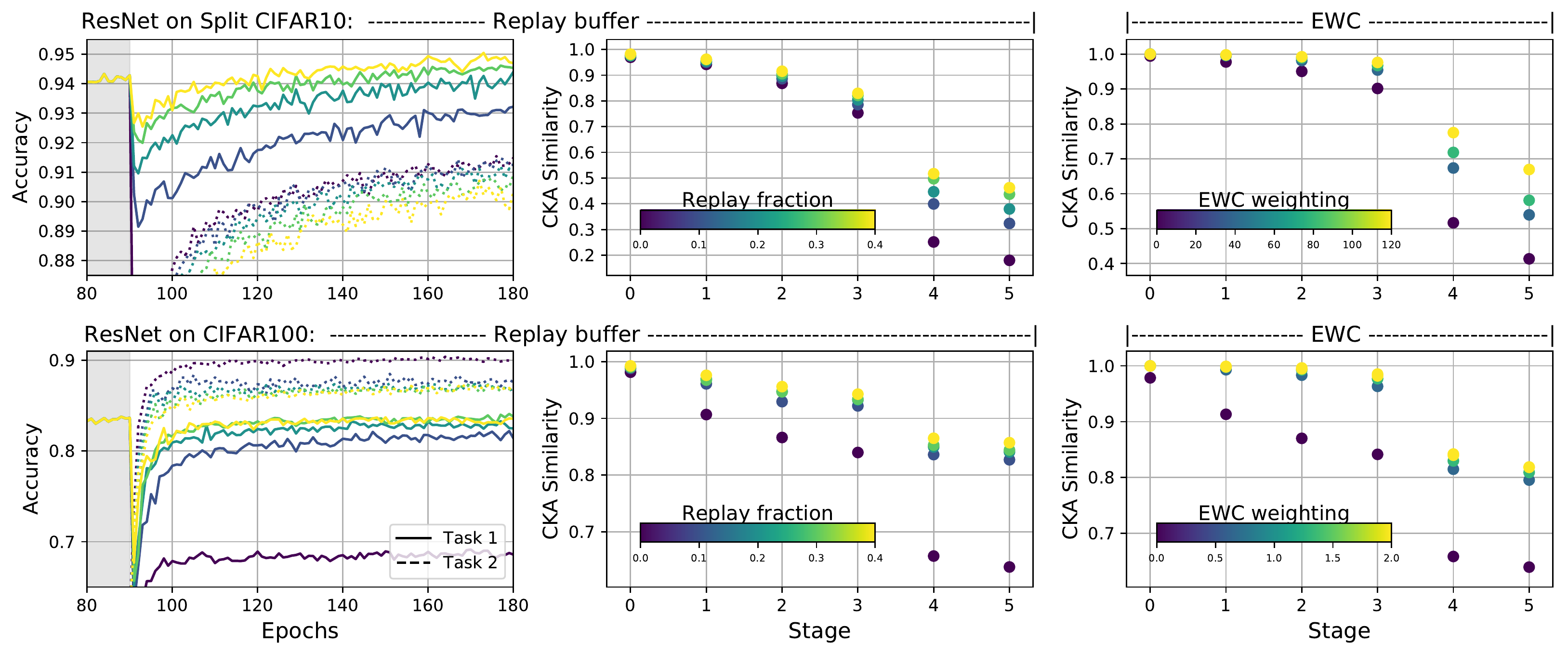}
\caption{\small \textbf{Both EWC and Replay Buffer mitigation strategies work by primarily reducing change in higher layer representations}. Using ResNet as a prototypical network architecture, we show the effects of these two popular mitigation strategies on both accuracies (left pane) and representations, as we vary the strength (fraction/weighting) of mitigation used (middle, right panes). These latter panes plot the same CKA comparisons as Figure \ref{fig-hiddenreps-cifar10}, showing much larger CKA scores (in particularly higher layers) with the lighter dots (strong mitigation) vs the purple dots (no mitigation.) Results for other architectures are in Appendix~\ref{sec:app_replayewc}.}
\label{fig-resnet-replay-buffer}
\end{figure}

As shown in Figure \ref{fig-resnet-replay-buffer}, our analysis finds that in both cases, the mitigation method's effectiveness stems from stabilizing higher layers during training on the second task. Concretely, we see that as we increase the strength of the mitigation strategy (through the magnitude of the quadratic penalty for EWC, and the replay fraction for the replay buffer) representations change less during the second task training (see Figure~\ref{fig-resnet-replay-buffer}), with the diminution particularly pronounced for deeper representations. 

\section{Semantics}
\label{sec:semantics}
A fundamental property of biological and artificial neural networks is that their predictive output is critically affected by semantic properties of their training data \cite{Saxe11537, MANDLER1993291}. In this section, building on the insights on catastrophic forgetting's effects on hidden representations, we delve into a deeper study of semantic task similarity and its effects on learned representations and forgetting.

We find catastrophic forgetting exhibits consistent behaviour with respect to task semantics, but our analysis also immediately reveals a semantic puzzle: in one setting, we observe (sequentially) similar tasks result in less forgetting, while in a different setting dissimilar tasks result in less forgetting. We demonstrate that this effect arises due to (i) task representations being a function of both the raw data instances \textit{and} the optimization process (ii) forgetting being most severe for tasks of \textit{intermediate} similarity. For this latter aspect, inspired by the findings in Section~\ref{sec-forgetting-hidden-reps}, we construct an analytic model (Section \ref{sec:analytic}) which makes mathematically precise the connection between maximal forgetting severity and intermediate task similarity, and empirically validate these findings. 

\subsection{Forgetting, Task Semantics and a Similarity Puzzle}
We begin by exploring the effect of task semantic similarities on forgetting, where we find a surprising puzzle across two experimental setups.
\label{sec:patterns}
\begin{figure}
     \centering
     \begin{subfigure}[b]{6.94cm}
         \centering
         \includegraphics[width=\textwidth]{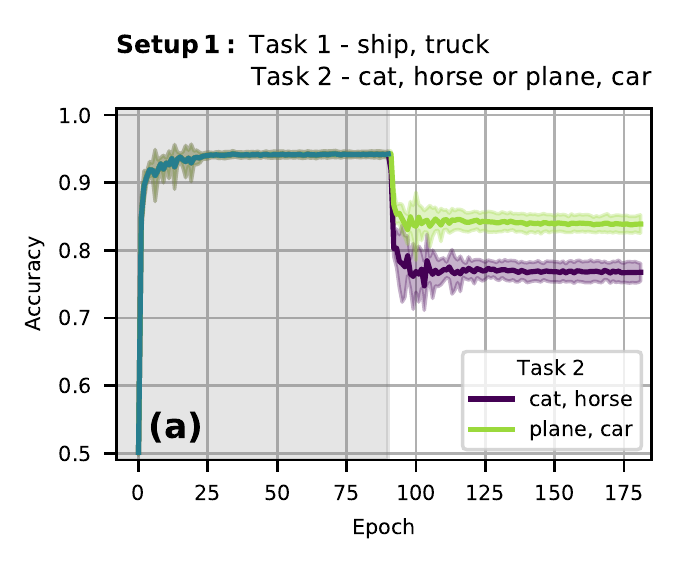}
         \phantomcaption{}
         \label{fig:2v2main}
     \end{subfigure}
     \hfill
     \begin{subfigure}[b]{6.94cm}
         \centering
         \includegraphics[width=\textwidth]{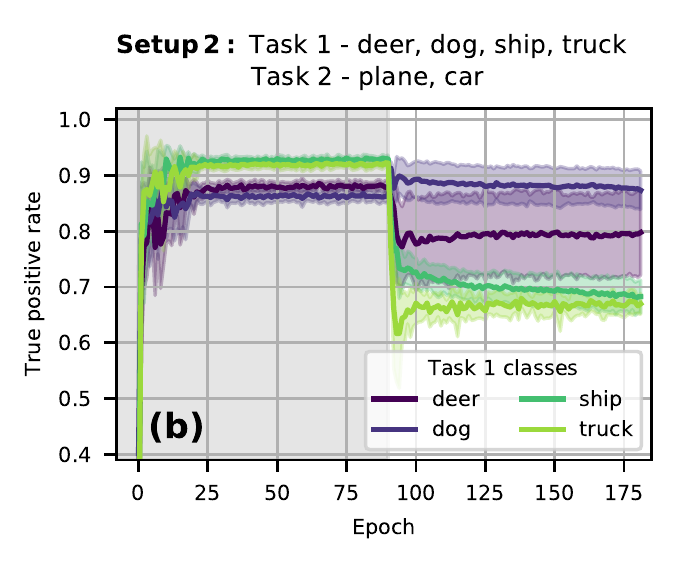}
         \phantomcaption{}
         \label{fig:4v2main}
     \end{subfigure}
        \vspace*{-8mm}\caption{\small \textbf{Forgetting severity consistently varies along semantic divisions.} (a) For Setup 1 a ResNet model trained initially on on object classification task forgets less on a second object task then an animal task. (b) For Setup 2 a four class model sees more forgetting in the object classes than the animal classes when trained on a second two class object task.
        }
        \label{fig:consistent_forgetting}
\end{figure}

\textbf{Setup 1} Our first set of semantic experiments consist of training on sequential binary classification task taken from CIFAR10. The two classes are chosen to either be two animals, such as cat versus horse, or two objects such as ship versus truck. We perform initial training on a given binary task, save the model weights, and compare forgetting between a second binary classification task consisting of animals and one consisting of objects (Figure~\ref{fig:2v2main} and Appendix~\ref{sec:app_semanticexp}). In this setup we find that semantically similar tasks lead to less forgetting.

\textbf{Setup 2} Our next set of experiments consist of an initial four category classification task with two animal and two object tasks followed by a binary classification task consisting of either two animals or two objects. In this setup, in contrast to Setup 1, we find that the dissimilar categories true positive fractions are hurt less then for the similar tasks (Figure~\ref{fig:4v2main} and Appendix~\ref{sec:app_semanticexp}).

Initially it seems as though these results are in tension. In the first setup similar tasks forget less while in the other they lead to more forgetting. Further investigation shows these results can be explained by two insights: (i) Forgetting is most severe for task representations of intermediate similarity. (ii) (Task) representational similarity is a function of the underlying data \emph{and} the optimization procedure.

In Setup 1, the initial task is either solely an animal or solely an object classification task. As a result, there is no pressure for the model to represent animals and objects differently. In this case all tasks are similar with the slightly less similar task causing more forgetting. In contrast for Setup 2, the four class classification task leads to representations which distinguish animals and objects and are different enough to cause diminished forgetting. We formalize this through an analytic model.

\subsection{An analytic model of forgetting semantics}
\label{sec:analytic}
To build an analytically tractable model of forgetting---called below the \textit{frozen feature model}---we write the network logits, $f(x)$, as a linear mapping applied to $p$ non-linear features, $g_{\mu}(w;x)$.
\es{ew:model_def}{
f(x)=\sum_{\mu=1}^{p}\theta_{\mu}g_{\mu}(w;x)\,.
}
Here we have split the model weights into final layer weights $\theta$ and additional intermediate weights $w$. We train this model sequentially on two tasks.  Inspired by our empirical observation in Section \ref{sec-forgetting-hidden-reps}, and Figures~\ref{fig-hiddenreps-cifar10}, \ref{fig-hiddenreps-cifar100} that forgetting is driven by deeper layers, we train the model ordinarily for the initial task, but approximate the features $g_{\mu}(\hat{w};x)$ as \textit{frozen} after first-task training, with only the final-layer weights $\theta_\mu$ allowed to evolve.

SGD training under the loss function $L(f,y)$ induces the following change in the output logits, $\Delta f_{t}$: 
\es{eq:sgd_change}{
\Delta f_{t}(x)=-\eta \sum_{x',y'\in \mathcal{D}_{\textrm{train}}^{(2)}}\Theta(x,x')\frac{\partial L(f(x'),y')}{\partial f}\,.
}
Here $\mathcal{D}_{\textrm{train}}^{(2)}$ is the second-task training data and $\Theta(x,x')=\sum_{\mu=1}^{p}g_{\mu}(\hat{w};x)g_{\mu}(\hat{w};x')$ is the data-data second moment matrix of the frozen features $g_{\mu}$. This kernel apears frequently in the analysis of linear models and wide neural networks (see Appendix~\ref{sec:app_analytic}) and measures the similarity between the features, $g_{\mu}$, on the new training data and a test point, $x$.

To study forgetting, we are interested in the evolution of the logits on the original task test data, $f_{t}(x)$ for $x\in X_{\textrm{test}}^{(1)}$. Informally, if the overlap $\Theta(x,x')$ is small, then the change in the logits is small and forgetting is minimal. If the representations were completely orthogonal, then the original task predictions would remain constant throughout all of the second task training and no forgetting would occur. This is summarized in the following lemma. 
\begin{lemma}
Let $f_{t}$ be the logits of a neural network trained via the frozen feature model. Let $x$ be an element of the initial task test set. Let us further denote the vector of the feature overlap aplied to the second task training set as $\vec{\Theta}(x):=\{\Theta(x,x'): x'\in X_{\textrm{train}}^{(2)}\}$ and the loss vector as $\vec{L}=\{L(f(x'),y'): x',y'\in \mathcal{D}_{\textrm{train}}^{(2)}\}$. With this, the change in the original task logits is bounded as
\es{eq:delta_bound}{
|\Delta f_{t}(x)|\leq \eta \left|\left|\vec{\Theta}\left(x \right)\right|\right|\left|\left|\frac{\partial \vec{L}}{\partial f}\right|\right|\,.
}
\end{lemma}
Thus in the frozen feature model representational similarity is necessary for forgetting to occur. It is not sufficient, however. As a simple counter example, consider the case where Task 1 and Task 2 data are identical, $\mathcal{D}_{\textrm{train}}^{(1)}=\mathcal{D}_{\textrm{train}}^{(2)}$. In this case the second task amounts to training the model head for longer on the first task data and thus does not lead to forgetting.

Thus we see that the frozen feature model provides a concrete example where sufficiently similar tasks, as measured through the feature overlap and target similarity, and sufficiently divergent tasks, as measured though the feature overlap, have minimal forgetting, while intermediate tasks forget. In Appendix~\ref{sec:app_analytic} we extend this to the multi-head setup. 
\begin{figure}
     \centering
     \begin{subfigure}[b]{6.94cm}
         \centering
         \includegraphics[width=\textwidth]{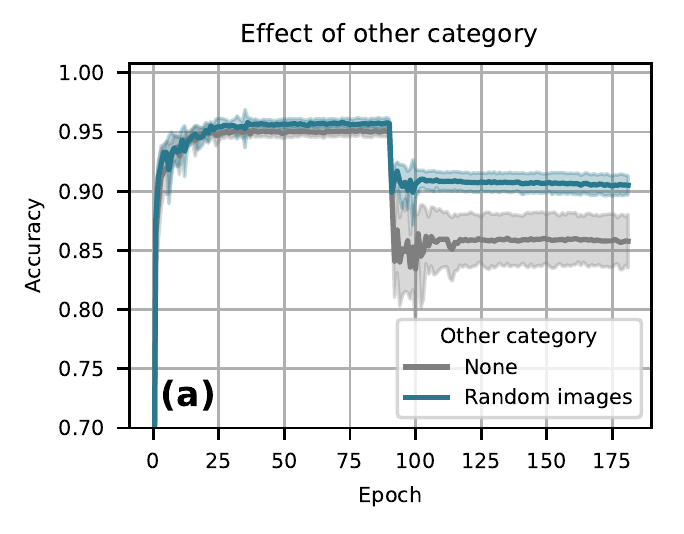}
        \phantomcaption{}
         \label{fig:orthomain}
     \end{subfigure}
     \hfill
     \begin{subfigure}[b]{6.94cm}
         \centering
         \includegraphics[width=\textwidth]{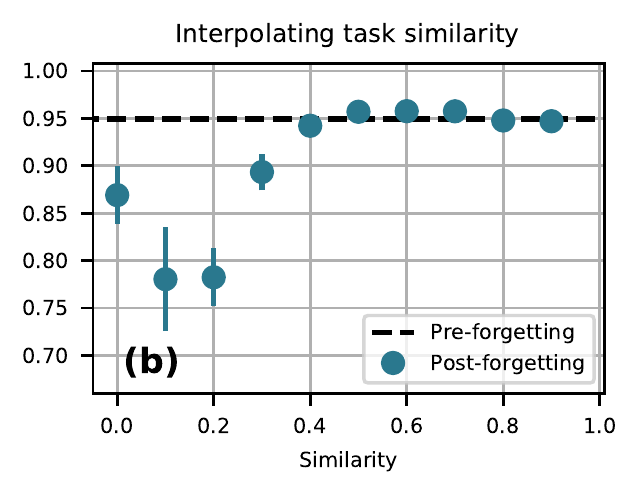}
         \phantomcaption{}
         \label{fig:interpmain}
     \end{subfigure}
         \vspace*{-8mm}\caption{\small \textbf{Task similarity affects performance for binary CIFAR-10 classification.} (a) We train an ordinary model (grey) and a model with a third \emph{other} category consisting of random images (teal) on sequential binary classification tasks. The other category encourages the model to represent new data as dissimilar and can soften the effect of forgetting. (b) We measure forgetting for sequential binary classification as we dial the degree of second task similarity by linearly interpolating between initial data and new data.}
        \label{fig:dialing}
\end{figure}
\subsection{Catastrophic Forgetting across varying Task Similarities}
Informed by the analytic model's insights, we next empirically study the degree of forgetting for different levels of task similarity.

\textbf{Increasing dissimilarity can help forgetting}
In Section~\ref{sec:patterns} we saw that the effect of forgetting can be diminished for dissimilar tasks, but only if the model representations register the tasks as dissimilar. For the four task setup of Figure~\ref{fig:4v2main}, this was accomplished automatically as a result of initially training on both objects and animals. Taking inspiration from this setup, during our initial binary task training, we introduce a third, \emph{other}, category consisting of images from all CIFAR-10 classes not included in the two training tasks to encourage the model to encode different images using dissimilar representations. In Figure~\ref{fig:orthomain} and Appendix~\ref{sec:app_semanticexp} we see that encouraging the model to represent the second task differently can lessen the degree to which the model forgets.  

\textbf{Interpolating similarity}
Our initial experiments in Figure~\ref{fig:consistent_forgetting} and analytic model in Section~\ref{sec:analytic} point towards a picture in which forgetting is maximized when the representations of an initial task and the subsequent task have intermediate similarity. To probe this phenomena further we emulated mixup \cite{zhang2017mixup}, a state of the art data-augmentation strategy which replaces the ordinary training images and labels with linear combinations of pairs of images and labels.

We first trained a model on a two class classification task consisting of either animals or objects with an additional other category to ensure that the model representations learned to distinguish animals from objects. We then dialed the second task dissimilarity by constructing a dataset, $\mathcal{D}_{\textrm{train}}^{(\lambda)}$, which was a linear mixture between a dissimilar second task and the initial task training data.
\es{eq:mixup_dataset}{
\mathcal{D}_{\textrm{train}}^{(\lambda)}=\{(\lambda x^{(2)}+(1-\lambda)x^{(1)}, \lambda y^{(2)}+(1-\lambda)y^{(1)}): (x^{(1)},y^{(1)})\in\mathcal{D}_{\textrm{train}}^{(1)},\,(x^{(2)},y^{(2)})\in\mathcal{D}_{\textrm{train}}^{(2)}\}\nonumber
}
By varying the mixing parameter, $\lambda$, we tuned the degree of task similarity. Figure~\ref{fig:interpmain} shows that the most extreme forgetting happens for intermediate similarity. 
\subsubsection{Semantics of forgetting in CIFAR100 distribution shift task}
\label{subsub:CIFAR-100Semantic}
\begin{figure}[h]
\begin{minipage}[l]{0.3\textwidth}
\includegraphics[width=6.5cm]{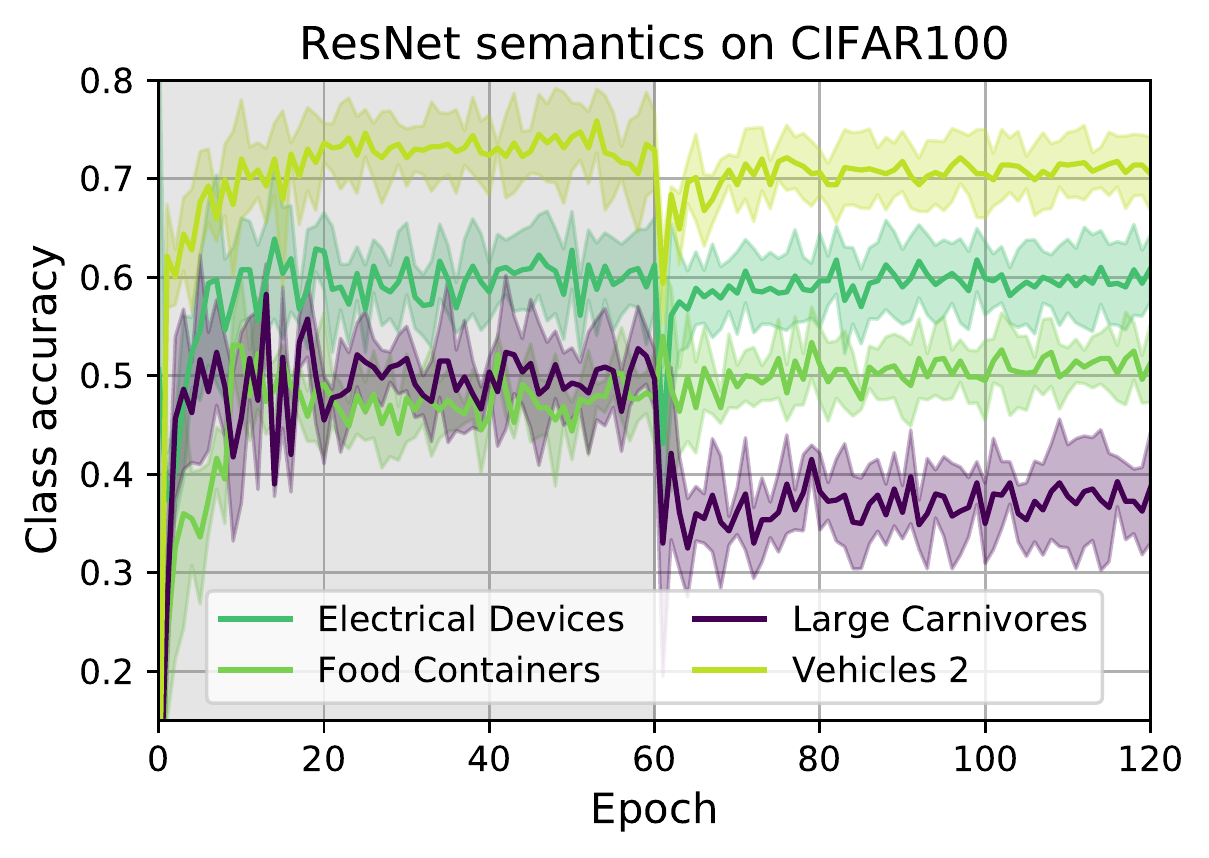}
\end{minipage}\hfill
\begin{minipage}[c]{0.5\textwidth}
\caption{\small \textbf{Semantics of forgetting in the CIFAR-100 distribution-shift setting}. We consider an initial task consisting of six super classes: \{Electrical Devices: clock, television; Food Containers: bottles, bowls; Large Carnivores: bear, leopard; Vehicles 2: lawn-mower, rocket; Medium Mammals: fox, porcupine; Small Mammals: hamster, mouse\}. For the subsequent task we change the mammal sub-classes to \{Medium Mammals: possum, raccoon; Small Mammals: shrew, squirrel\}. We observe maximal forgetting for the remaining animal super-class Large Carnivores.}\label{fig:semanticscifar100}
\end{minipage}
\end{figure}

 To investigate the role of task similarity in our CIFAR-100 distribution shift setup we consider a classification problem distinguishing between six of the CIFAR-100 super-classes. We sequentially train on two collections of sub-tasks with a semantically clear distribution shift. For example in Figure~\ref{fig:semanticscifar100} we train on three animal and three object super-classes, \{Electrical Devices, Food Containers, Large Carnivores, Vehicles 2, Medium Mammals, Small Mammals\}. We then change the underlying distribution of the Medium Mammals and Small Mammals super-classes by altering the sub-classes.
Of the remaining four tasks we observe that the semantically similar animal category, Large Carnivores, suffers the most. Again we find evidence that sufficiently dissimilar representations suffer less forgetting. Further empirical experiments supporting this pattern can be found in Appendix~\ref{sec:app_semanticexp}.
\section{Conclusion}
In this paper, we have studied some of the fundamental properties of catastrophic forgetting, answering important open questions on how it arises and interacts with hidden representations, mitigation strategies and task semantics. By using representational similarity techniques, we demonstrated that forgetting is not evenly distributed throughout the deep learning model but concentrated at the higher layers.
Consistent with this (and despite other mitigation possibilities such as orthogonalizing representations), we find that diverse forgetting mitigation methods such as EWC and replay buffers primarily act to stabilize higher layers. Looking at the effects of task semantic similarities, we find forgetting behavior shows an important dependence on the optimization process as well as the training data, with intermediate similarity amongst tasks leading to maximal forgetting. We formalize this insight through an analytic model and additional empirical experiments systematically varying task similarity. Overall, the results of our work both provide a foundation for a deeper, more principled understanding of catastrophic forgetting, as well as promising future directions to help its mitigation across the many ways it arises in practice.

\nocite{caliban2020github}

\section*{Acknowledgements}
The authors gratefully acknowledge valuable conversations with Samy Bengio, Ekin Dogus Cubuk, Orhan Firat, Guy Gur-Ari, Boris Hanin, Aitor Lewkowycz, Behnam Neyshabur, Samuel Ritchie, and Ambrose Slone.  Additionally, we thank the authors of the image classification library at \url{https://github.com/hysts/pytorch_image_classification}, on top of which we built much of our codebase.

\bibliography{refs}
\bibliographystyle{unsrtnat}

\appendix

\clearpage

\section{CIFAR-100 Distribution Shift Task}
\label{sec-SuppCIFAR-100}
Here we provide a concrete example of the CIFAR-100 distribution-shift task.  As mentioned in the main text, in this task, the model must identify, for CIFAR-100 images, which of the 20 superclasses the image belongs to.  The difference in tasks comes from the difference in subclasses which make up the superclass.  As an example, we take the five superclasses \textit{aquatic mammals}, \textit{fruits and vegetables}, \textit{household electrical devices}, \textit{trees}, and \textit{vehicles-1}\footnote{The CIFAR-100 dataset features two vehicle superclasses, denoted \textit{vehicles-1} and \textit{vehicles-2}}, with the corresponding task 1 subclasses (1) dolphin, (2) apple, (3) lamp, (4) maple tree, and (5) bicycle and task 2 subclasses (1) whale, (2) orange, (3) television, (4) willow, and (5) motorcycle.  A key feature of this setting is that task-specific components (including multiple heads) are precluded since the model does not know the task identity either at inference or time.  While we do not explore it here, this setup also allows for continuously varying the data distribution, another situation likely to occur in practice.  
\section{Methods}\label{sec:app_methods}
\subsection{Models}

Here we detail the model architectures used in this study.  We use standard versions of the popular VGG~\cite{simonyan2014deep}, ResNet~\cite{he2015deep}, and DenseNet~\cite{huang2016densely} architectures. These architectures have varying structural properties (e.g., presence or absence of skip connections), meaning observed behaviors common to all of them are likely to be robust and generalizable.

\textbf{VGG}:
The VGG model we use consists of five stages.  Each stage comprises a convolutional layer, a ReLU nonlinearity, another convolutional layer, another ReLU nonlinearity, and finally a MaxPool layer.  Each convolution uses a 3-by-3 kernel with unit stride and padding.  The MaxPool operation uses a 2-by-2 kernel with stride 2.  The number of channels by stage is 16, 32, 64, 128, 128.  We do not use batch normalization in the VGG.

We test two different version of the VGG: one with just the stages described above, and another with two fully connected layers, of width 1024, after the convolutional layers.

\textbf{ResNet}:  Our ResNet consists of a initial 3-by-3 convolutional layer with 32 channels, followed by a 2d batch norm operation.  This initial pair is followed by four stages, each consisting of two residual blocks per stage.  Each block consists of two conv-BN-ReLU sequences, with a shortcut layer directly routing the input to add to the final ReLU preactivations.  All convolutions are 3-by-3, with unit padding.  In all but the first block, the first convolutional layer downsamples, with a stride of 2; the second convolutional layer maintains stride of 1.  The number of channels doubles each block, starting from a base level of 32 channels in the first block.

\textbf{DenseNet}:  Like ResNet, our DenseNet consists of an initial 3-by-3 convolution, followed by four dense blocks.  In between each pair of dense blocks is a transition block.  Following the final dense block is a batch-normalization, ReLU, and average pooling operation to generate the final features.  Our DenseNet is characterized by a growth rate of 12 and compression rate of 0.5.  The first stage of the DenseNet features 6 blocks, with the number of blocks doubling in each subsequent stage. 

\subsection{Training}

All networks are trained using cross-entropy loss with SGD with momentum ($\beta$ = 0.9), using a batch size 128.  We do not use learning-rate schedules here, leaving that investigation to future work.  To better correspond with practical situations, however, we choose a learning rate and total number of epochs such that interpolation (of the training set) occurs.  For the split CIFAR-10 task, this typically corresponds to training for 30 epochs per task with a learning rate of 0.01 (VGG), 0.03 (ResNet), and 0.03 (DenseNet).  We use weight decay with strength 1e-4, and do not apply any data augmentation.  For the split CIFAR-100 task, we usually train for 60 or 90 epochs per task, with other the other hyperparameters identical to the CIFAR-10 case.  

For multi-head networks (used in split CIFAR-10), each head  is initialized with weights drawn from a normal distribution with variance $1/n_f$, where $n_f$ is the number of features; biases are initialized to zero.  We do not copy the parameters from the old head to the new when switching tasks.  

For the split CIFAR-10 setup, the plots shown in the main text refer to experiments in which the initial task was classifying between the five categories \emph{airplane}, \emph{automobile}, \emph{bird}, \emph{cat}, and \emph{deer}; and the second task comprising \emph{dog}, \emph{frog}, \emph{horse}, \emph{ship}, and \emph{truck}.  To verify that our results were not unique to the particular split of CIFAR-10 we chose, we used three other splits:  \emph{airplane}, \emph{bird}, \emph{deer}, \emph{frog}, \emph{ship} (task 1) and \emph{automobile}, \emph{cat}, \emph{dog}, \emph{horse}, \emph{truck} (task 2); \emph{automobile}, \emph{cat}, \emph{dog}, \emph{horse}, \emph{truck} (task 1) and  \emph{airplane}, \emph{bird}, \emph{deer}, \emph{frog}, \emph{ship} (task 2); and \emph{dog}, \emph{frog}, \emph{horse}, \emph{ship}, \emph{truck} (task 1) and \emph{airplane}, \emph{automobile}, \emph{bird}, \emph{cat}, \emph{deer} (task 2). 

\subsection{Representational Similarity and Centered Kernel Alignment}
To understand properties of neural network hidden representations, we turn to representational similarity algorithms. A key challenge in analyzing neural network hidden representations is the lack of \textit{alignment} --- no natural correspondence between hidden neurons across different neural networks. Representational similarity algorithms propose different ways to overcome this --- one of the first such algorithms, SVCCA \cite{raghu2017svcca, morcos2018insights}, uses a Canonical Correlation Analysis (CCA) step to align neurons (enable invariance) through linear transformations.

We use centered kernel alignment (CKA), proposed in \cite{kornblith2019similarity}, to measure the similarity between two representations of the same dataset which is invariant to orthogonal transformation and isotropic scaling (but not arbitrary linear transformations).  Given a dataset of $m$ examples, we compare two representations $X$ and $Y$ of that dataset (say, from two different neural networks or from the same neural network with different parameters), with $n_x$ and $n_y$ features respectively; that is, $X \in \mathbb{R}^{m\times n_x}$ and $Y \in \mathbb{R}^{m\times n_y}$.  Then the linear-kernel CKA similarity between the two representations is given by 
\begin{equation}
    \mathrm{CKA}(X,Y) = \frac{||X^T Y||^2_F}{||X^T X||^2_F ||Y^T Y||^2_F}
\end{equation}

\subsection{Elastic Weight Consolidation}\label{SuppEWC}

Developed by Kirkpatrick et al. in 2017, elastic weight consolidation (EWC) adds a term to the loss function of the new task to get the regularized loss function $\mathcal{L}(\vec{\theta})$:  
\begin{equation}
    \mathcal{L} (\vec{\theta}) = \mathcal{L}_B (\vec{\theta}) + \frac{\lambda}{2}\sum_i F_i\cdot \left(\theta_i - \theta_{A,i}^* \right)^2.
\end{equation}
Here, $\vec{\theta}$ are the model parameters; $\mathcal{L}_B(\vec{\theta})$ is the unregularized loss function of the new task; $\lambda$ is a parameter which controls the strength of the regularization; $F_i$ is the diagonal of the Fisher information matrix with respect to the parameters; and $\vec{\theta}^*_A$ are the model parameters after having been trained on task A.  We compute the Fisher information via using the squared gradients of the log-likelihood, averaged over a subset of the trainset.  

Following standard practice, in our experiments we estimate the Fisher information using a subset of the full training data; for all of our models on CIFAR-10, 200 samples proved sufficient to be reasonably confident of a converged estimate. We do not include a delay between the start of second-task training and the application of the EWC penalty.  Naturally, we do not apply the EWC penalty to the parameters of the head in the multi-head setting.

\section{Forgetting versus network width}

\begin{figure}[h]
\centering
\includegraphics[width=13.5cm]{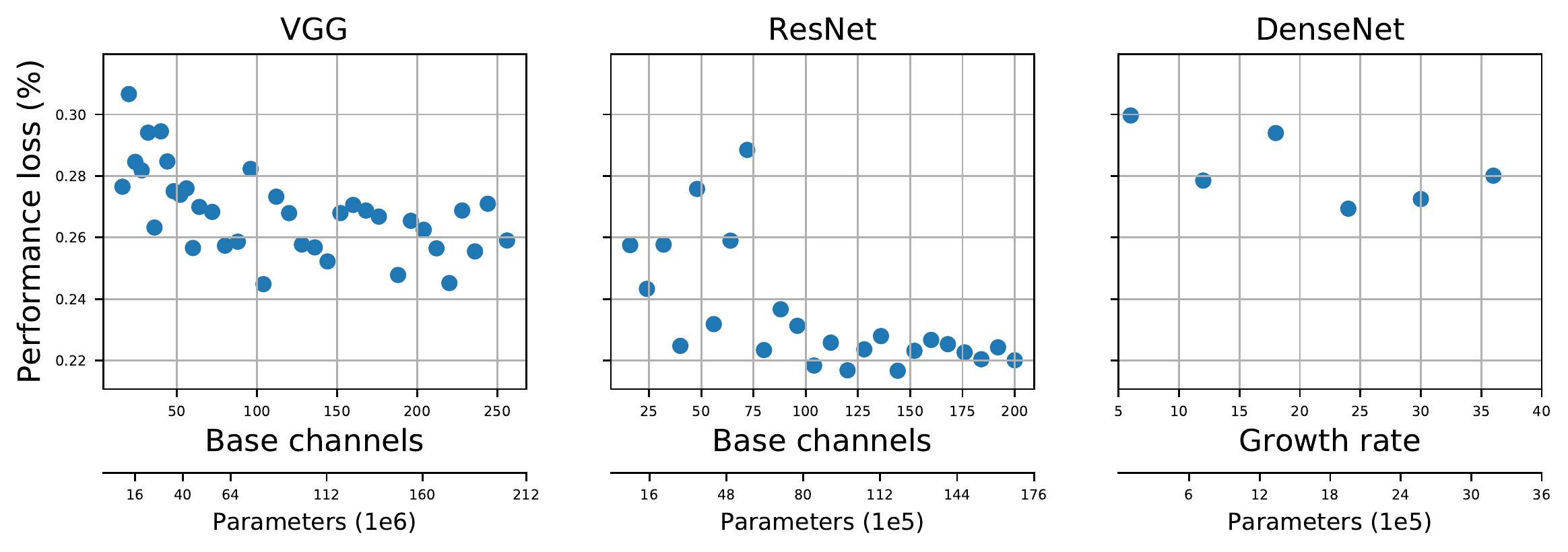}
\caption{\textbf{Forgetting as a function of network width}. (a) VGG, (b) ResNet, and (c) DenseNet networks of varying widths were trained on two subsets of CIFAR-10 classes.  We trained each of these (multi-head) networks for 30 epochs on each task with constant $\eta$, enough to achieve perfect training accuracy.  Further training details can be found in the text.  From the plots, which show the percent drop in accuracy (on task 1) due to training on task 2, the effect of width on the performance drop seems to be minimal, even sweeping across roughly a factor of ten in the number of model parameters.}
\end{figure}

Inspired by recent work~\cite{gilboa2019wider} showing that wider networks can learn better features than their narrower counterparts, we sought to determine whether wider networks also exhibit less catastrophic forgetting than narrower ones.  We tested forgetting versus width for VGG, DenseNet, and ResNet networks on the split CIFAR-10 task in the multi-head setting.  For VGG and ResNet, the width was varied by changing the number of channels in the initial convolutional layer and scaling the rest of the network layers concurrently while maintaining the network shape; for DenseNet, the width was varied by by changing the growth rate.  

Surprisingly, though we found that the performances of the networks on each task improved as they grew wider, the amount of forgetting in each network, measured by the percent drop in accuracy (on task 1) due to training on task, exhibited a minimal dependence on network width.

\section{Headfirst training}

\begin{figure}[h!]
\begin{center}
\includegraphics[width=13.5cm]{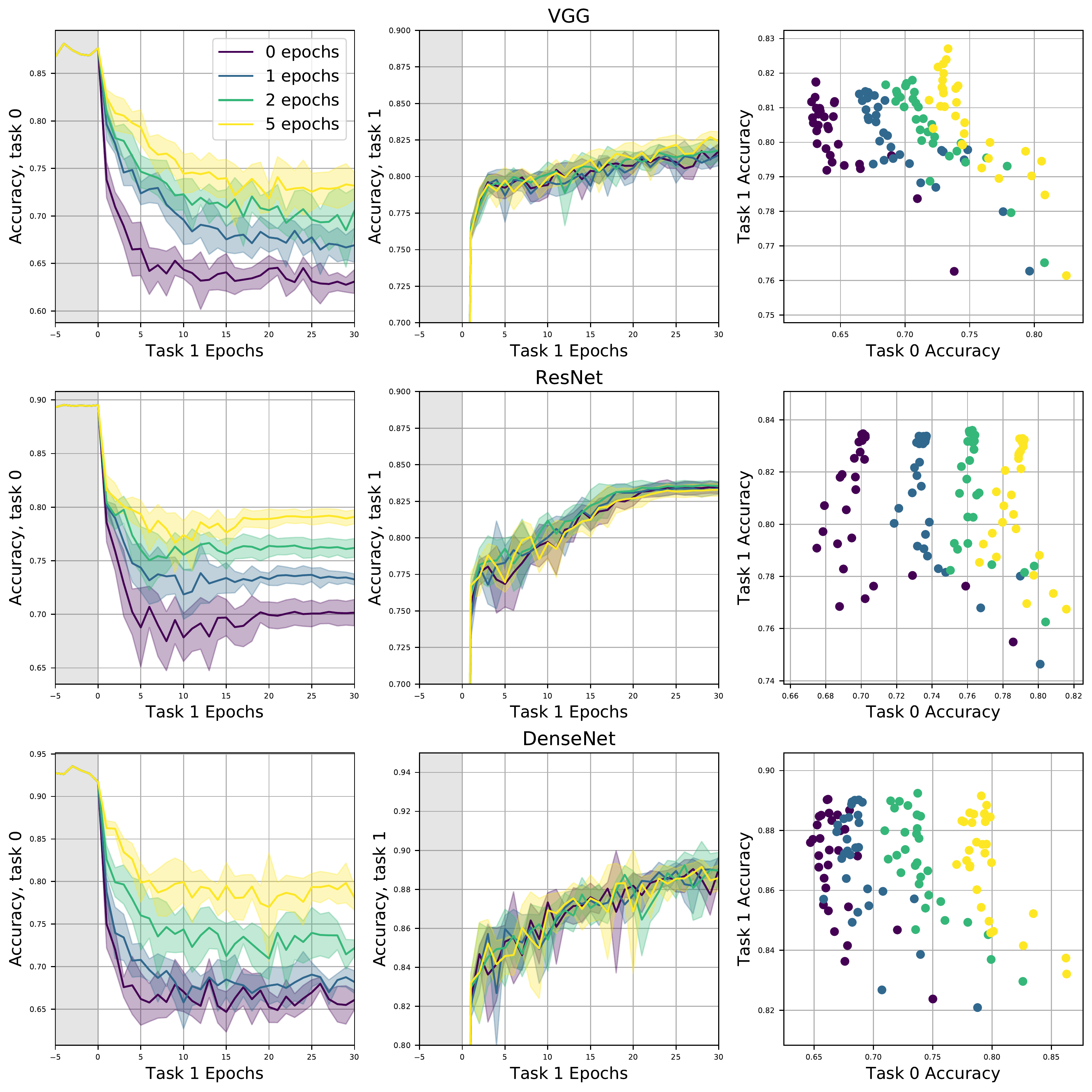}
\caption{\textbf{Effect of training readout layer separately between task switches}.  (a) VGG, (b) ResNet, and (c) DenseNet models were trained on two subsets of CIFAR-10 classes:  task 0 (\emph{dog}, \emph{frog}, \emph{horse}, \emph{ship}, and \emph{truck}), and task 1 (\emph{airplane}, \emph{automobile}, \emph{bird}, \emph{cat}, \emph{deer}).  When switching tasks, we first trained \emph{only} the final classification layer (known as the \emph{readout layer} or \emph{head}) for a specified number of epochs (here $0$, $1$, $2$, and $5$) from a random Gaussian initialization.  During this period, the rest of the model parameters are held fixed; after the head-only training, the entire model is trained as usual.  Across all architectures, training only the head consistently improves the forgetting performance of the model (its accuracy on task 0) while having a negligible impact on the performance on the next task (task 1).  }
\label{fig:headfirst}
\end{center}
\end{figure}

When training multi-head networks on sequences of tasks, we found that the networks suffered a smaller performance drop on the original task, i.e. forgot less, if \emph{only} the new task head was trained for a few epochs prior to training the full network.  In Figure~\ref{fig:headfirst}, this effect is shown for all three network architectures on the split CIFAR-10 task.  Training only the head for up to five epochs seemed to improve the original-task performance without sacrificing performance on the new task.  This suggests that coadaptation between the freshly-initialized head and the rest of the model contributes to forgetting when there is no headfirst training.   

\section{Task-specific stages}

Our observations that forgetting is driven primarily by the higher layers suggests that in a setting with known task identities (i.e. in the split CIFAR-10 setting described in the main text but not the CIFAR-100 distribution-shift setting), making the deeper stages of the network task-specific would likely mitigate a large portion of forgetting on the old tasks while still allowing for full performance on the new tasks.  In particular, since the changes in deeper represntations are much greater than their shallower counterparts, only a few task-specific stages should be necessary to get significant performance increases on the old task.  For each of the three network architectuers we considered (VGG, ResNet, and DenseNet), we investigated the performance of these task-specific networks on the split CIFAR task.  We intend these results to be further evidence of the role of deeper layers in forgetting, not a proposed mitigation scheme.   

\begin{figure}[h]
\centering
\includegraphics[width=13.5cm]{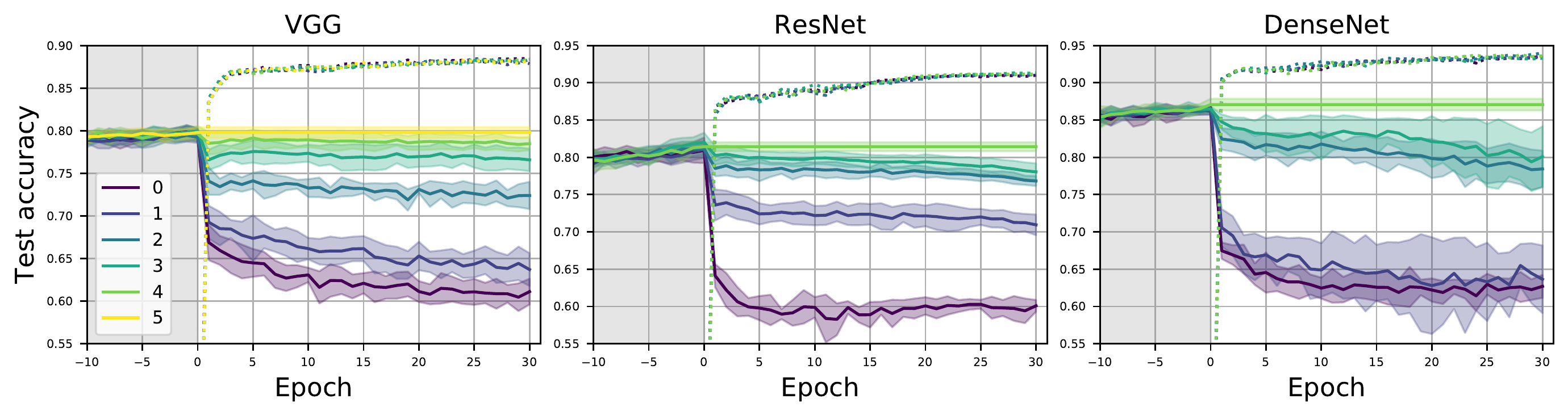}
\caption{\textbf{Forgetting in networks with task-specific stages on CIFAR-10}.  Using the same settings as Figure 1 of the main text, we run networks with task-specific layers on split CIFAR-10. We make deeper layers task-specific before shallower layers.   Performance increases with number of task-specific layers, denoted by the line color (see legend).  In all architectures, two task-specific stages is sufficient to recover significant performance. \label{fig:ynet}}
\end{figure}

Results, shown in figure~\ref{fig:ynet}, show that for all architectures, performance dramatically rises with the number of task-specific stages.  For all architectures, having the deepest two stages be task-specific recovers a significant fraction of the forgotten task performance.

\section{Additional replay buffer and EWC results}\label{sec:app_replayewc}

This section gives CKA similarity measurements to accompany those in the main text presented in figure~\ref{fig-resnet-replay-buffer}.  The main text figure showed results for ResNet on split CIFAR-10 using both EWC and a replay buffer; here we give results for the other two architectures we investigated, VGG and DenseNet, also on split CIFAR-10.  Additionally we show replay buffer results for split CIFAR-100.  These plots show that the results discussed in the main text apply broadly across architectures and datasets.

\section{Layer reset and retrain experiments}\label{sec:app_resetretrain}

As an additional probe of the degree to which catastrophic forgetting impacts the representations of various network layers, we perform a reset-and-retrain experiment on the split CIFAR-10 task, using the same training setup as in Figure~\ref{fig-hiddenreps-cifar10} of the main text.  The reset-and-retrain experiment is designed to probe the degree to which network representations get corrupted and can no longer be used by a network of the same architecture to productively classify images.  To investigate this, after having trained the network on Task 2, we freeze parameters of stages $1$ through $N$ to their post-task-2 values, reset the parameters of stages $N+1$ onwards to their post-task-1 values, and then retrain those parameters on the Task 1 loss function.  This is a natural, operational measure of the degree to which layer $1$ through $N$ representations have been corrupted.  

\begin{figure}[h]
\centering
\includegraphics[width=14cm]{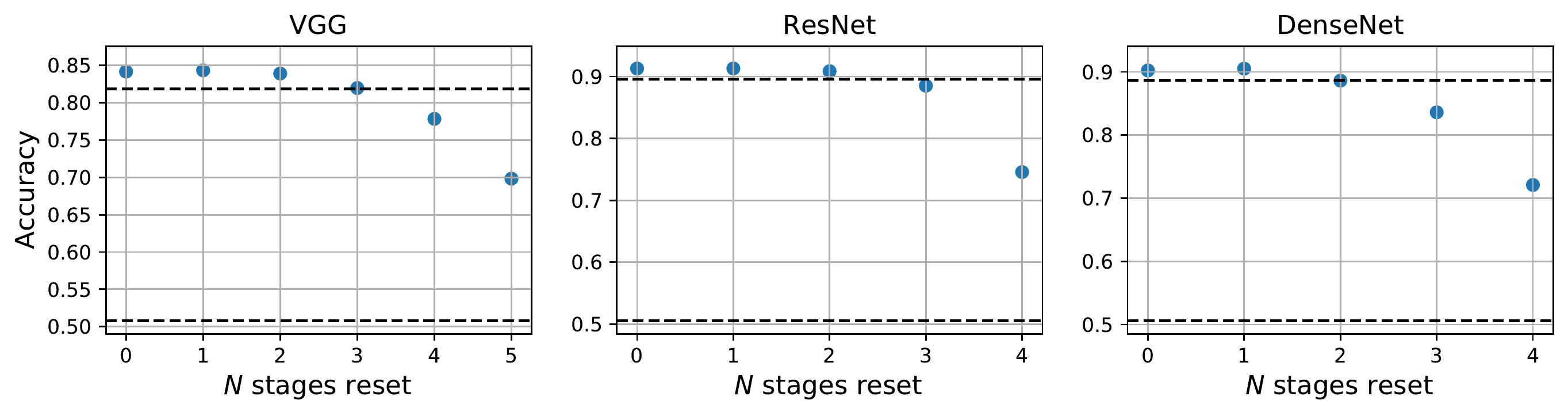}
\caption{\textbf{Stage reset and retrain experiments.} Consistent with the observations in the freeze-training, layer reset, and CKA experiments, resetting and retraining only a few stages from the top is enough to recover the performance before forgetting.  \label{fig:reset_retrain}}
\end{figure}

The results of this experiment, shown in Figure~\ref{fig:reset_retrain}, tell a similar story to that of Figure~\ref{fig-hiddenreps-cifar10} of the main text.  In particular, we see even retraining the network with all but the last couple of layers frozen recovers nearly the pre-forgetting performance of the network.  

\begin{figure}[h]
\centering
\includegraphics[width=14cm]{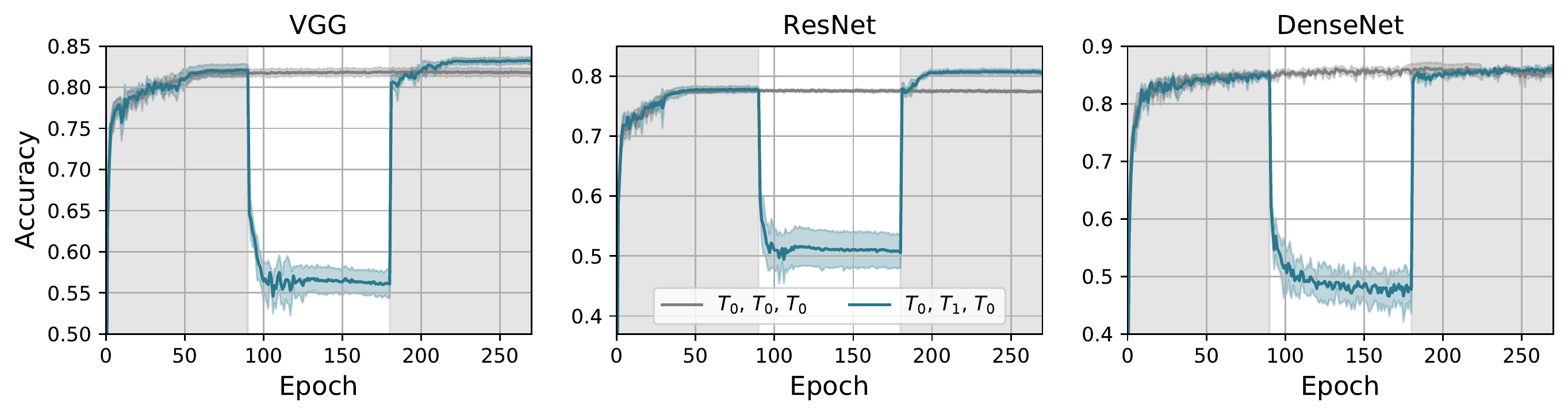}
\caption{\textbf{Training on a different task can improve performance on the original task}. Networks are trained on two five-class splits of CIFAR-10.  The gray curve shows the test accuracy on task 0 during training on solely that task for 270 epochs;  the teal curve shows the task-0 test accuracy when training on task 0 for 90 epochs, then on task 1 for 90 epochs, and task 0 again for 90 epochs.  While we see a sharp performance drop for the duration of training on task 1, the performance is quickly recovered once we begin training on task 0, and in fact surpasses the performance of the model only exposed to task-0 data.  This effect is consistent across different splits of the original dataset.  Of the architectures we studied, this effect is most pronounced for ResNet, but also exists to some degree for VGG and DenseNet models. \label{fig:repeated}}
\end{figure}

Some of the performances in Figure~\ref{fig:reset_retrain} even outperform the pre-forgetting network performance, suggesting that network representations of intermediate layers may actually be improved by training on Task 2.  Further suggestive results are shown in Figure~\ref{fig:repeated}, in which we retrain the full network again on Task 1 after the training on Task 2.  For ResNet and, to some degree, VGG, training on the second task clearly improves performance on the first.  

\section{Linear regression on forgotten network activations}\label{sec:app_linreg}

To further probe the manner in which catastrophic forgetting affects the internal representation of networks, we perform the following experiment (in the split CIFAR-10 setting).  We train a (multi-head) model on the first task for some time, resulting in model $M_1$, and then train on the second task for some time, resulting in model $M_2$.  Due to forgetting, the model $M_2$ of course performs significantly worse than does $M_1$ on the first task.  But, to probe how much information has been lost internally, we train a linear model to classify between the first-task categories using only the internal activations of $M_2$ on the first-task images.  We measure both the amount of performance which we can recover using this linear regression, and the weight placed by the linear model on various stages of the network.

Model performances are given in Table~\ref{tab:linear}.  For all three architectures, the performance of the model before forgetting is naturally the highest, with a performance drop of up to forty percent due to forgetting.  Remarkably, however, training a linear model on the internal post-forgetting activations recovers a substantial fraction of the performance loss: for ResNet and VGG, the linear model on post-forgetting activations is only a percent less accurate than the model before forgetting.  As a control, we also train a linear model on the activations of networks at initialization (denoted the \emph{random lift} in the table).  Because the number of activations is much higher than the number of pixels in a CIFAR-10 image, this control is necessary to ensure that the boost in performance we see is not simply due to performing a high-dimensional lift which renders the data linearly separable.  As the table shows, however, this random lift performance is quite poor.

\begin{figure}[h!]
\centering
\includegraphics[width=14cm]{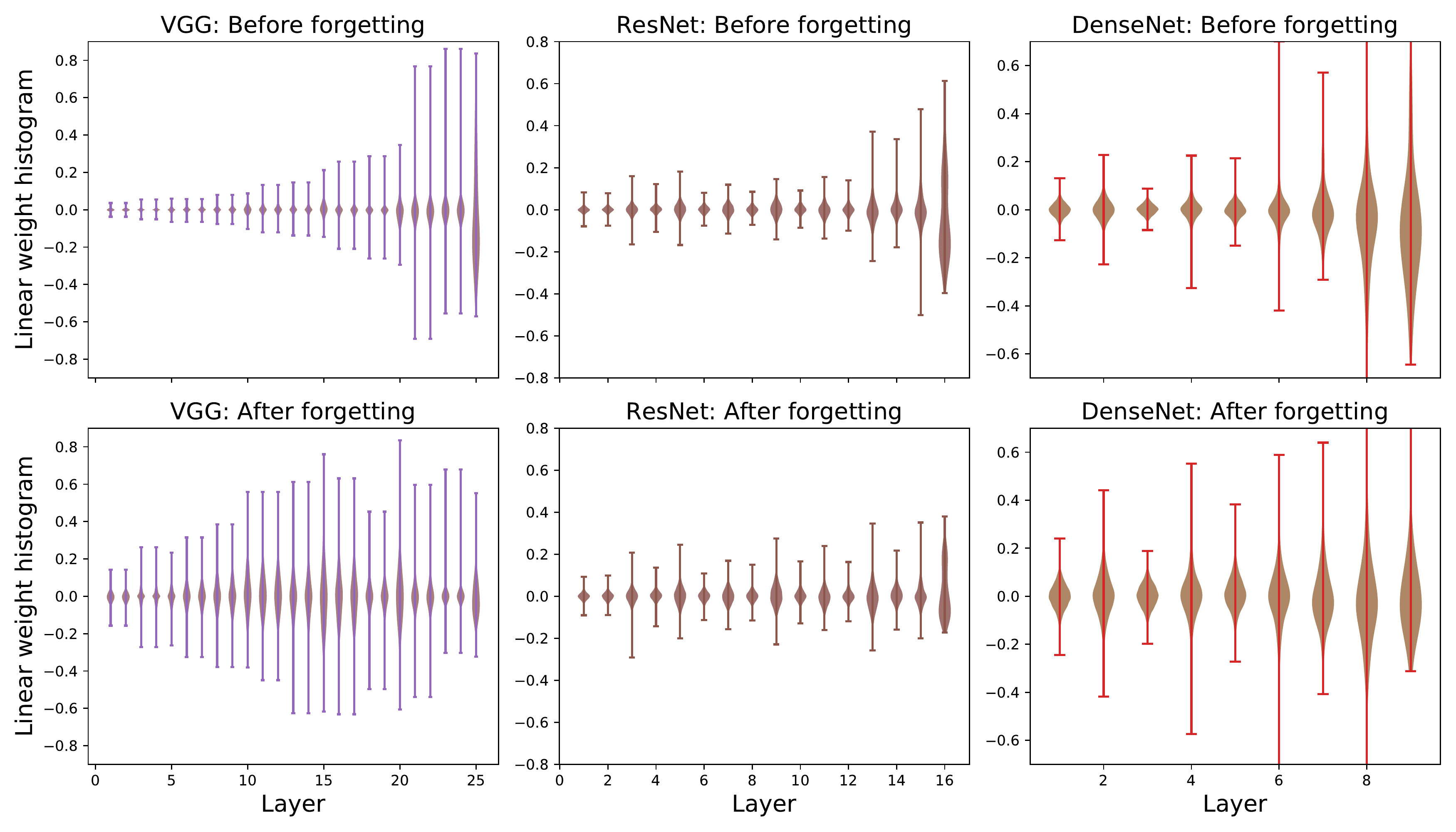}
\caption{\textbf{Weights of linear models trained on internal activations.} \label{fig:linear_weights}}
\end{figure}

Figure~\ref{fig:linear_weights} shows the weights placed by the linear model on the various internal activations of the network.  We use the activations both before and after forgetting.  Before forgetting, a linear model trained on the activations places most importance on the final layers, as expected; however, after forgetting, the model learns to use information stored in earlier-layer representations of the network.  The reduced weighting of the final layers is consistent with the experiments we describe in the main text showing that deeper representations suffer the most from forgetting.

\begin{table}
\centering
 \begin{tabular}{||c c c c||} 
 \hline
  & DenseNet & ResNet & VGG \\ [0.5ex] 
 \hline\hline
 Pre-forgetting accuracy & 0.869 & 0.7714 & 0.7970 \\ 
 \hline
Post-forgetting accuracy & 0.4754 & 0.5924 & 0.5877 \\
 \hline
 Linear regression on activations & 0.802 & 0.7592 & 0.7896 \\
 \hline
 Linear regression on random lift & 0.447 & 0.4198 & 0.543 \\ 
 \hline
\end{tabular}
\vspace{2mm}
\caption{\textbf{Performance of linear model trained on internal activations}.}
\label{tab:linear}

\end{table}

\section{Additional CIFAR-100 Distribution shift experiments}\label{sec:app_CIFAR-100exprmts}

\begin{figure}[h!]
\centering
\includegraphics[width=14cm]{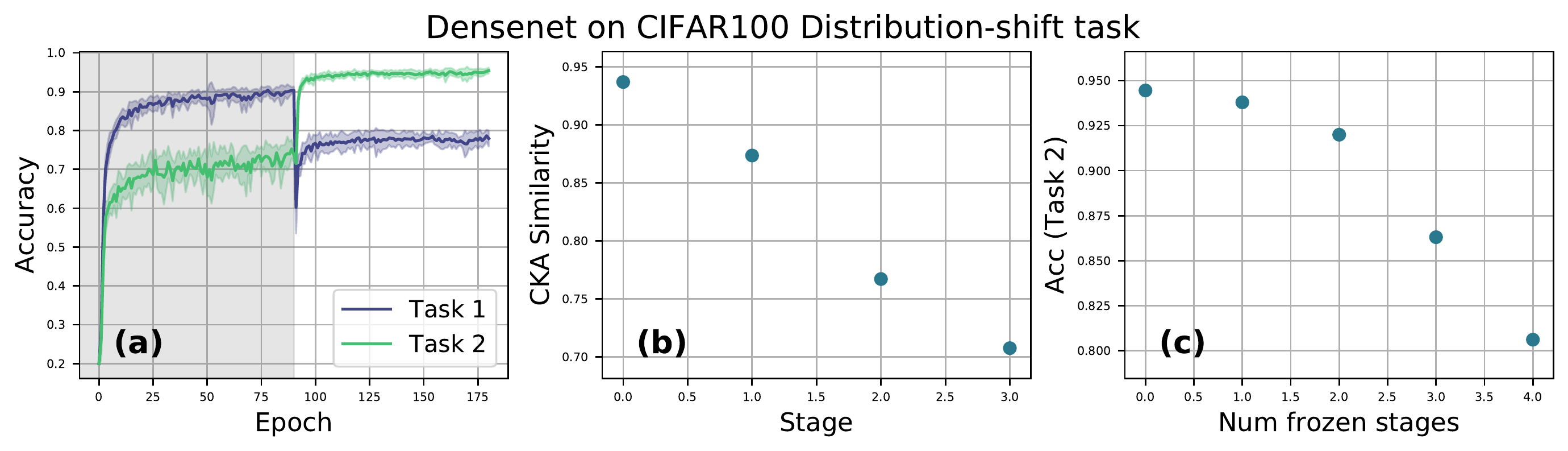}
\caption{\textbf{Change in Densenet representations under CIFAR-100 distribution shift}.\label{fig:DenseNetC00shift}}
\end{figure}

\begin{figure}[h!]
\centering
\includegraphics[width=14cm]{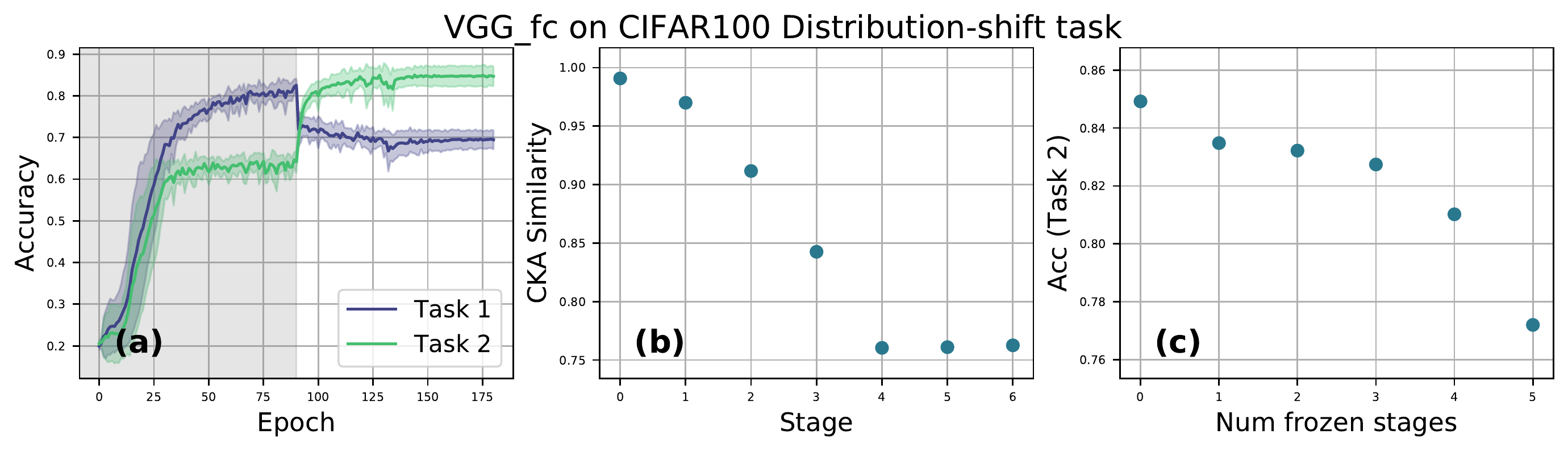}
\caption{\textbf{Change in VGG representations under CIFAR-100 distribution shift}.\label{fig:VGGC100shift}}
\end{figure}

Here we show the change in representations due to training on distribution-shifted CIFAR-100 for the VGG (Figure~\ref{fig:DenseNetC00shift}) and DenseNet (Figure~\ref{fig:VGGC100shift}) networks, to accompany the ResNet results in the main text.  

\section{Additional semantic experiments}\label{sec:app_semanticexp}
\begin{figure}[h!]
     \centering
     \includegraphics[width=\textwidth]{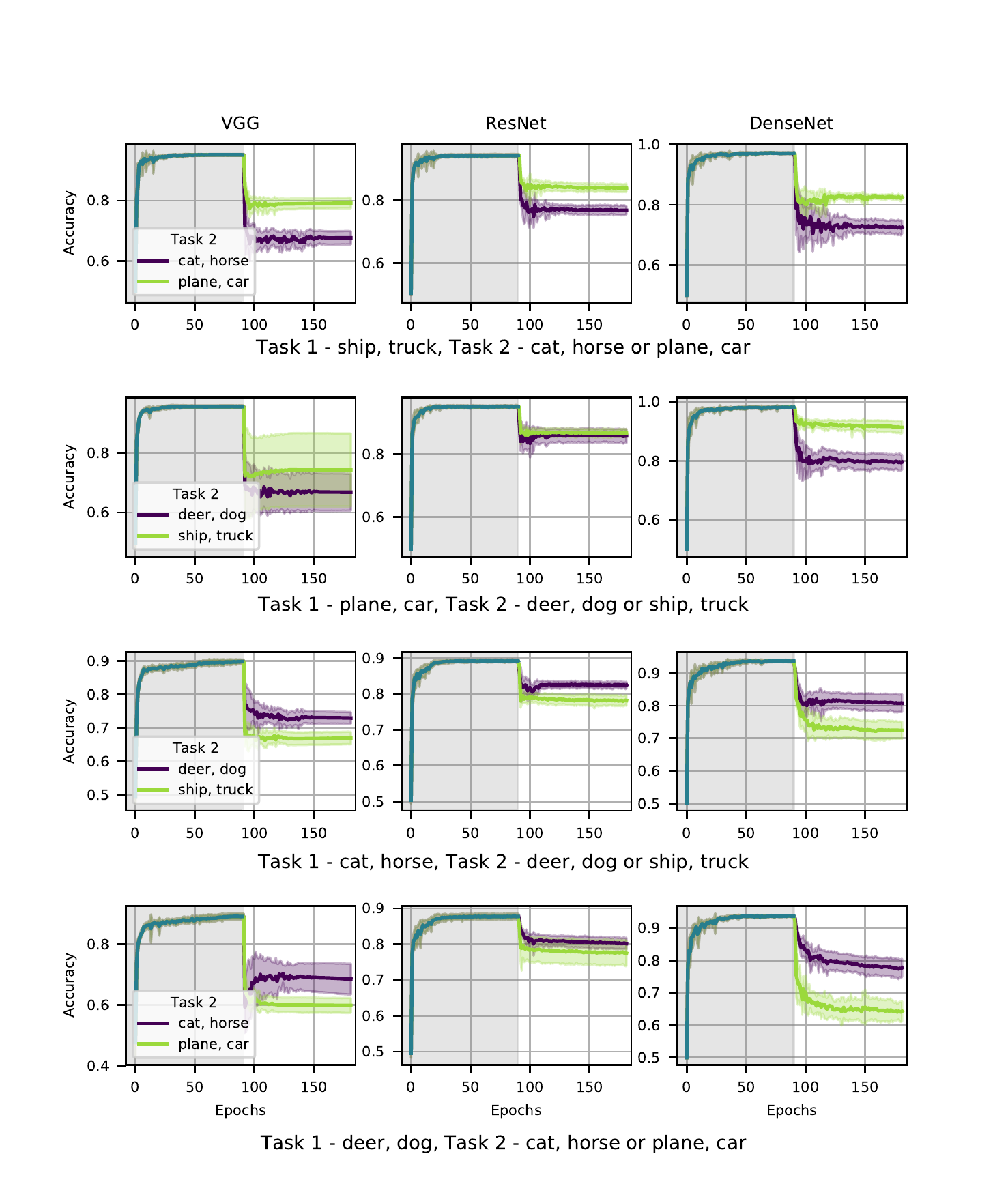}
   \caption{\small \textbf{Sequential binary classification tasks show consistent semantic structure.} We train on sequential binary subsets of CIFAR-10, distinguishing between two animals or two objects. We find that less forgetting occurs when the initial task is more similar to the second task.  
        }
        \label{fig:app_2v2}
\end{figure}

\begin{figure}[h!]
     \centering
     \includegraphics[width=\textwidth]{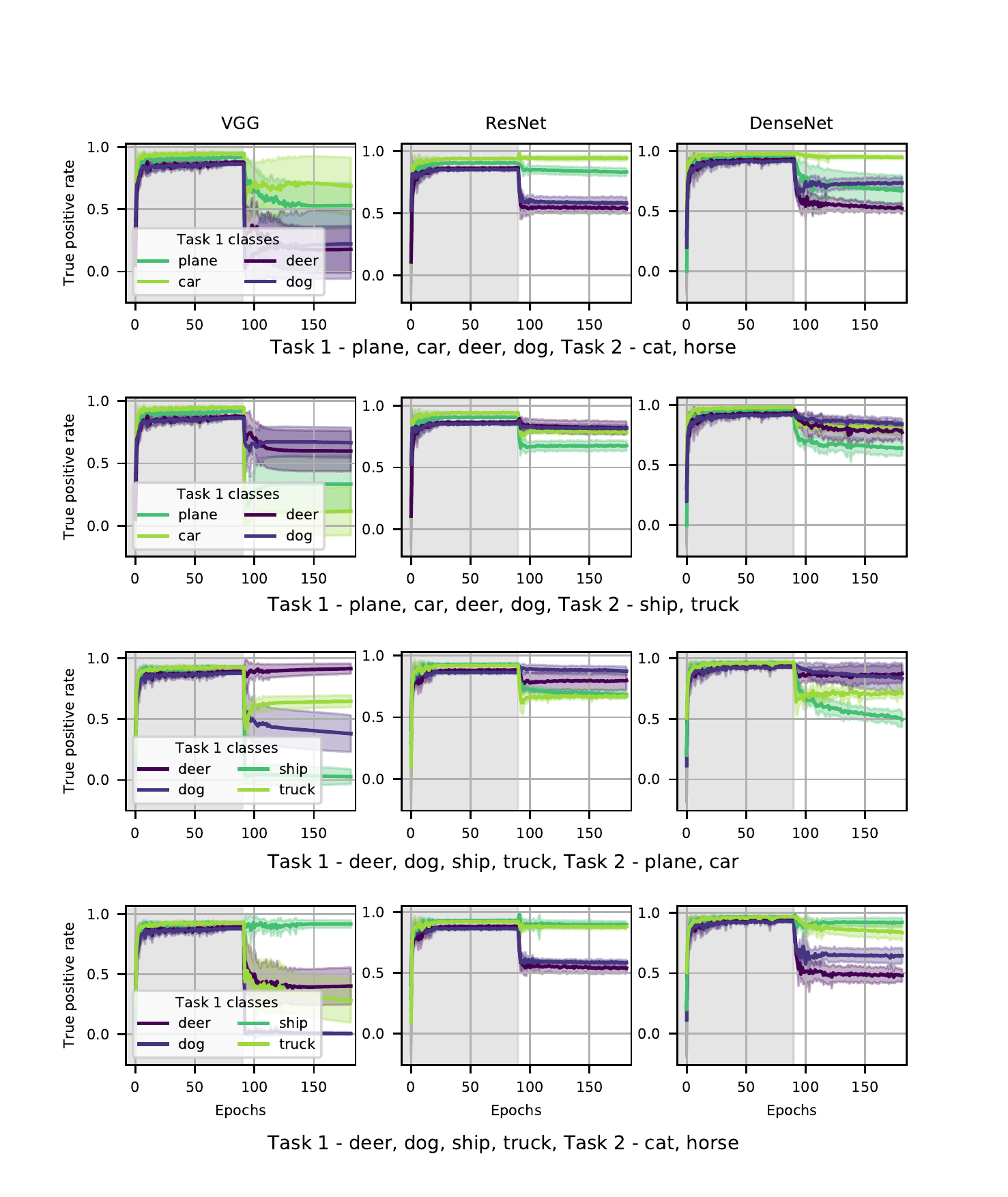}
   \caption{\small \textbf{Sequential four class and binary tasks show decreased forgetting for dissimilar tasks.} We train on sequential subsets of CIFAR-10, initially distinguishing between four classes (two animals and two objects) and then distinguishing between either two animals or two objects. We find that less forgetting occurs for categories that are dissimilar to those used in the second task.}
        \label{fig:app_4v2}
\end{figure}

\begin{figure}[ht]
\begin{subfigure}{.5\textwidth}
\centering
\includegraphics[width=.9\linewidth]{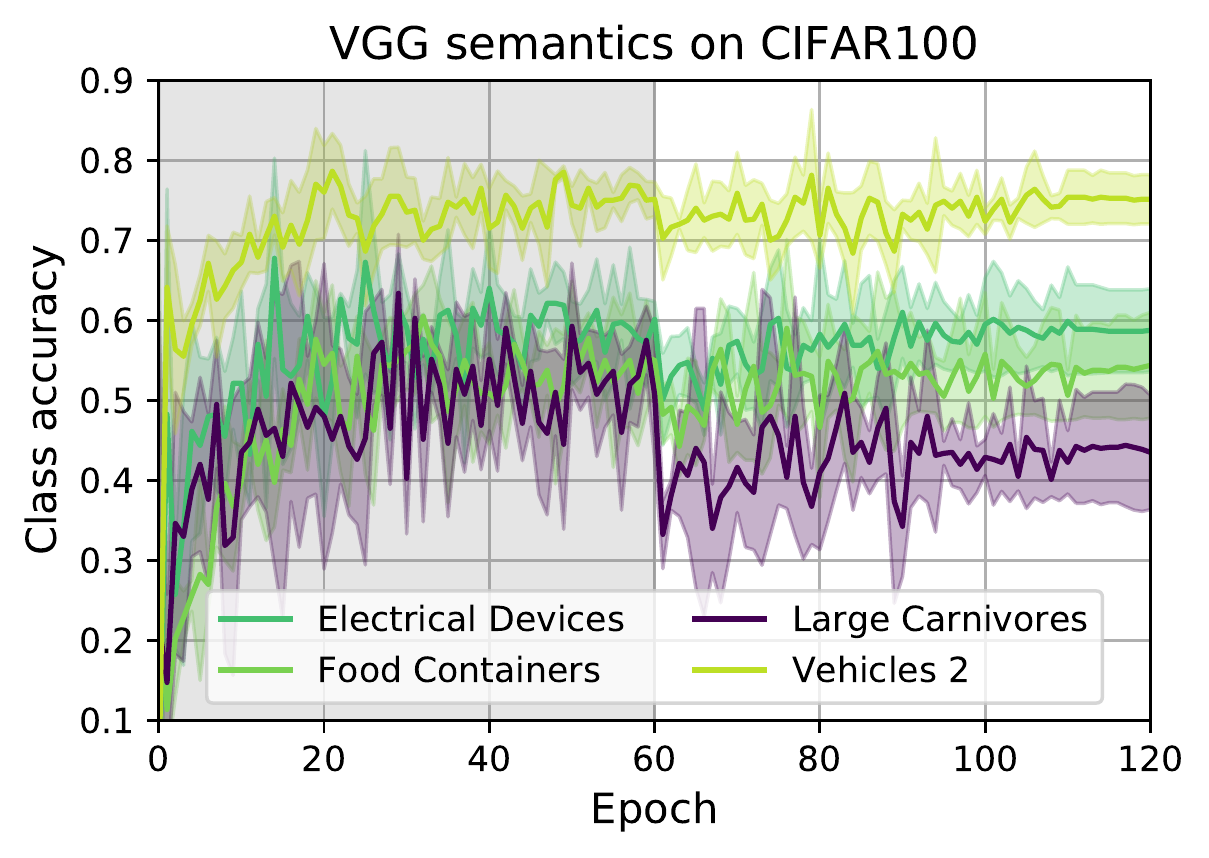}
\end{subfigure}
\begin{subfigure}{.5\textwidth}
\centering
\includegraphics[width=.9\linewidth]{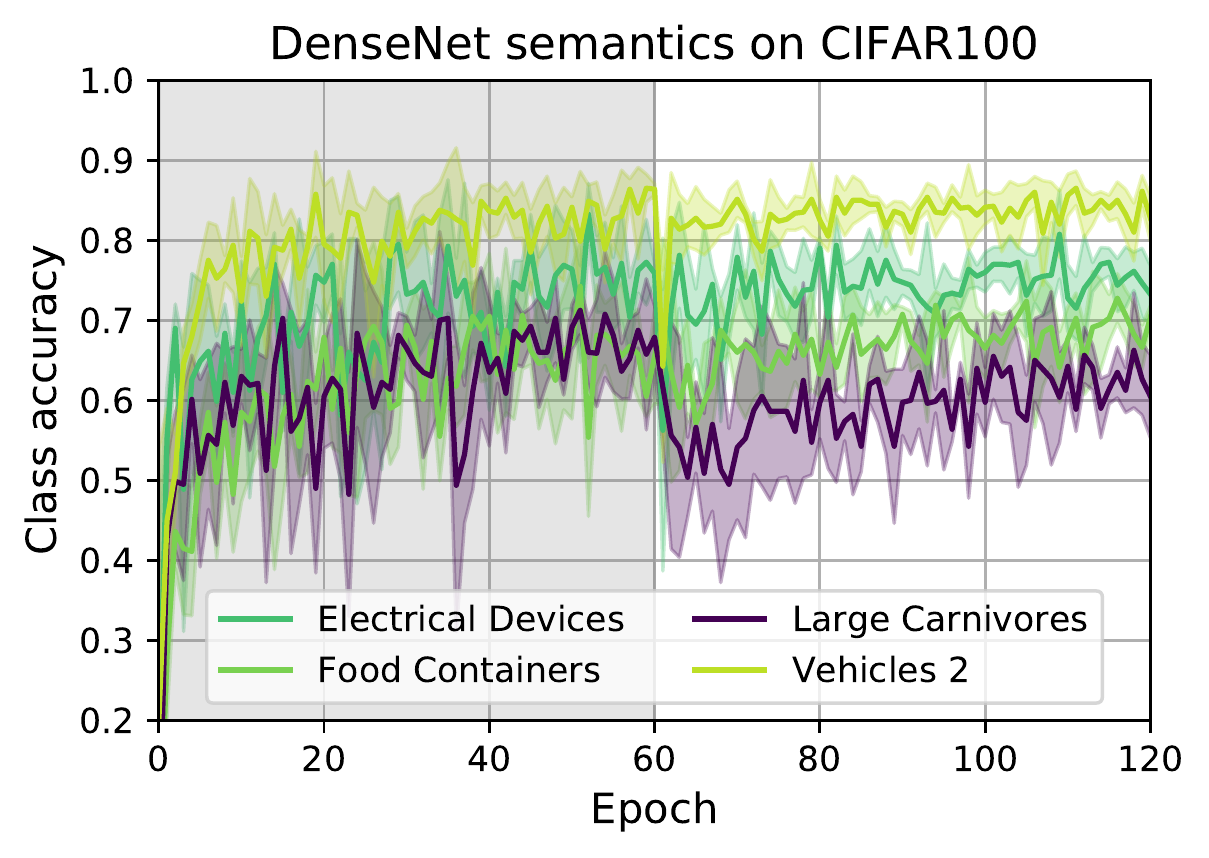}
\end{subfigure}
\caption{\textbf{Semantic structure of forgetting on the CIFAR-100 distribution-shfit task for VGG and DenseNet models}}
\label{fig:vggc100semantic}
\end{figure}

In this section we present additional realizations of the semantic experiments presented in Section~\ref{subsub:CIFAR-100Semantic}. In particular we present results for sequential binary classification tasks (Setup 1, main Figure~\ref{fig:2v2main} and Figure~\ref{fig:app_2v2}), four class classification followed by two class classification (Setup 2, main Figure~\ref{fig:4v2main} and Figure~\ref{fig:app_4v2}), and the CIFAR-100 distribution shift task (main Figure~\ref{fig:semanticscifar100} and Figure~\ref{fig:vggc100semantic}). We find consestant results accross different choices of objects and animals and across VGG, ResNet, and DenseNet architectures. In particular we find that when models are encouraged to distinguish objects and animals dissimilar categories lead to less forgetting (Setup 2 and CIFAR-100 setup). In contrast when models are trained with only a single semantic category (Setup 1), there is no pressure to distinguish objects from animals and similar categories lead to less forgetting. We observe one exception to the intuitive semantic groupings of categories. For VGG in Setup 2 (Figure~\ref{fig:app_4v2}) the truck category behaves similarly to animal classes in that the performance suffers more when the second task is animal classification then when it is object classification.

\section{Analytic model}\label{sec:app_analytic}
Here we consider extensions and applications of the analytic model introduced in Section~\ref{sec:analytic}. Both the single and multi-head models relate the change in predictions during training on a second task to the overlap between features, $\Theta(x,x')=\sum_{\mu}g_{\mu}(x)g_{\mu}(x')$, evaluated on the initial and second task data.\footnote{This feature overlap matrix shows up frequently when studying linear models, where it is sometimes known as a reproducing kernel, or wide neural networks, where it is called the neural tangent kernel \cite{Jacot2018NeuralTK} and governs SGD evolution.}

\paragraph{Multi-head model}
In the main text we focused for simplicity on an analytic model of single head forgetting. Here we construct a solvable model for the multi-head setup. We consider a model consisting of non-linear features $g(w;x)$, a linear layer with weights, $\theta$, and a read out head $h^{(i)}$.
\es{eq:multi_model_def}{
f^{(i)}(x)=\sum_{a,\mu}h^{(i)}_{a}\theta_{a\mu}g_{\mu}(w;x)\,.
}
Here $\mu$ runs over the feature dimensions, while $a$ runs over the output dimensions of our additional linear layer. For the initial task we use the head $h^{(1)}$, while for the second task, we swap the head and use $h^{(2)}$.

Again we consider a model trained without restrictions on Task 1 using head $h^{(1)}$. For Task 2, we first swap the head and then perform head only training on head $h^{(2)}$. After training the head, we freeze the head, $h^{(2)}$, and the features $g(w;x)=g(\hat{w};x)$, where $\hat{w}$ is the value of the feature weights after training on the initial task. This model is again inspired by our observation that forgetting is driven by changes in the latter network layers. The multi head model is additionally inspired by our observing that after head first training, the CKA similarity between the second task head before and after second task training is relatively high (see Table~\ref{tab:cka_multihead}). 

After training the second task head, we continue training the weights $\theta$. The model output evolves as
\es{eq:multi_model_ouput}{
\Delta f^{(i)}_{t}(x)=-\eta \sum_{x',y'\in \mathcal{D}_{\textrm{train}}^{(2)}}\Theta(x,x')\frac{\partial L(f^{(j)}(x'),y')}{\partial f}h^{T(j)} h^{(i)}\,.
}
Again we find that if the representation overlap matrix $\Theta(x,x')=\sum_{\mu}g_{\mu}(\hat{w};x)g_{\mu}(\hat{w};x')$ is small between the features evaluated on the initial task and second task data then the predictions do not change significantly. If this overlap is zero, then the predictions are constant. In the multi-head setup we are considering here, we see that the change in predictions is further proportional to the similarity between the model heads, $h^{T(j)} h^{(i)}$. 

Finally, we note that we have considered a final linear layer before the readout head for simplicity. One can instead include a ReLU non-linearity without changing the essential point, that the change in model output is governed by the overlap matrix, $\Theta$. 

\begin{table}[]
\begin{center}
\begin{tabular}{l|c|c|c}
                            & VGG               & ResNet            & DenseNet            \\
                            \hline
No headfirst training       & 0.147 $\pm$ 0.051 & 0.183 $\pm$ 0.001 & 0.162 $\pm$ 0.008  \\
5 epochs headfirst training & 0.550 $\pm$ 0.029 & 0.673 $\pm$ 0.010 & 0.700 $\pm$ 0.012 \\
\hline
\end{tabular}
\vspace{3mm}
\caption{\textbf{Headfirst CKA similarity}. CKA similarity between the readout layer after headfirst training and after final training on the second task.}
\label{tab:cka_multihead}
\end{center}
\end{table}

\begin{figure}
     \centering
     \begin{subfigure}[b]{6.94cm}
         \centering
         \includegraphics[width=\textwidth]{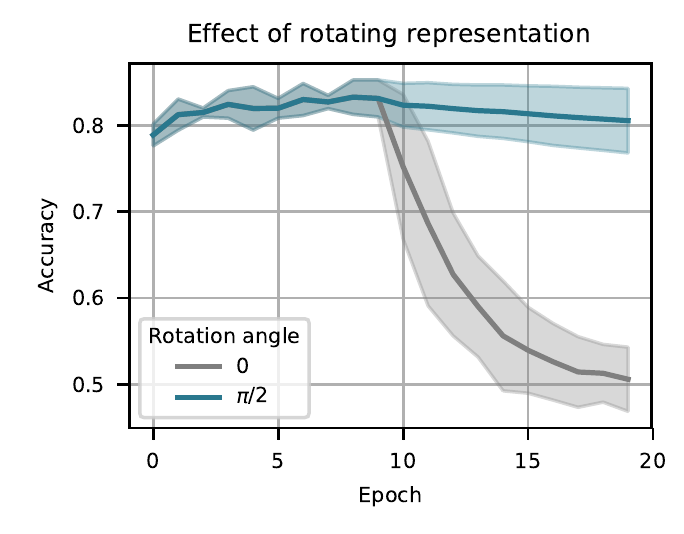}
         \label{fig:orthosupp}
     \end{subfigure}
     \hfill
     \begin{subfigure}[b]{6.94cm}
         \centering
         \includegraphics[width=\textwidth]{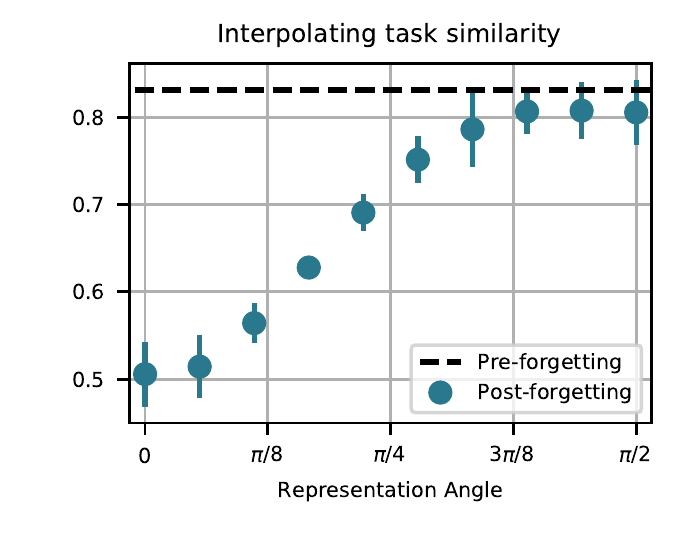}
         \label{fig:interpsupn}
     \end{subfigure}
         \vspace*{-8mm}
    \caption{\textbf{Rotating representations shows diminished forgetting for dissimilar tasks}. Here we consider the performance of our frozen feature model on sequential binary CIFAR-10 tasks (Task 1: car vs plane, Task 2: cat vs horse). We use a model built from a two hidden-layer fully connected network with ReLU activations. We increase decrease representational similarity by explicitly rotating the second task features and find as predicted that this leads to diminished forgetting.}\label{fig:dialingsup}
\end{figure}

\paragraph{Rotating representations in the analytic model.}
We can use our analytic model (Section~\ref{sec:analytic}) to investigate how forgetting depends on representation similarity. We again consider sequential binary tasks (car vs plane) followed by (cat vs horse). We then tune the overlap of our initial task and final task features by explicitly rotating the frozen features for the second task to produce new features $g'(\theta;\hat{w};x)$.
\es{eq:rotated_features}{
g'(\theta;\hat{w};x) = R(\theta)g(\hat{w};x)\,.
}
Here, if we have $P$ features, $R(\theta)\in\mathbb{R}^{P\times P}$ is a one-parameter family of rotation matrices designed such that $R(0)=\mathbf{1}$ and $g^{T}(\hat{w};x)g'(\pi/2;\hat{w};x')$ for $x\in X^{(1)}_{\textrm{test}}$ and $x'\in X^{(2)}_{\textrm{train}}$ is small. Explicitly, we take
\es{eq:rotationdef}{
R(\theta)=V^{T}\prod_{i=1}^{\lfloor{P/2}\rfloor}r_{i,P-i}(\theta)V\,,
}
where $r_{ij}(\theta)$ is the two dimensional rotation matrix between axes $i$ and $j$, and $V$ is the orthogonal matrix appearing in the singular value decomposition of the feature by data matrix on the initial task, $g(X^{(1)}_{\textrm{test}})=UDV$. We find that explicitly enforcing task dissimilarity via a large rotation minimizes the effect of forgetting (Figure~\ref{fig:dialingsup}).

\begin{figure}
     \centering
     \begin{subfigure}[b]{6.94cm}
         \centering
         \includegraphics[width=\textwidth]{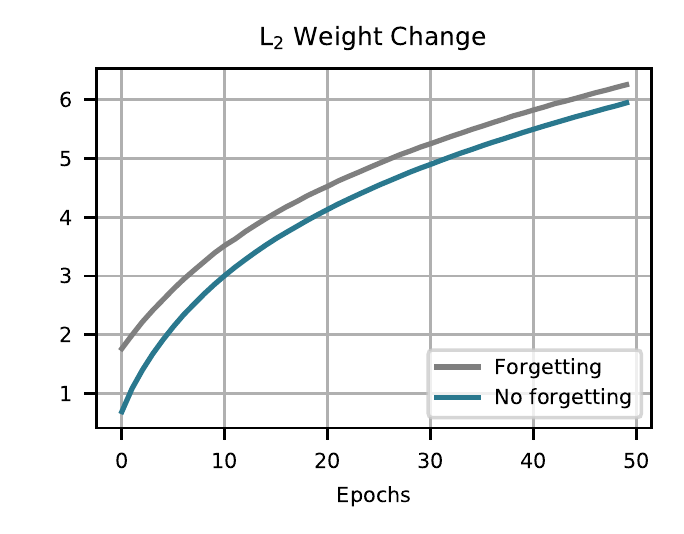}
         \label{fig:l2dist}
     \end{subfigure}
     \hfill
     \begin{subfigure}[b]{6.94cm}
         \centering
         \includegraphics[width=\textwidth]{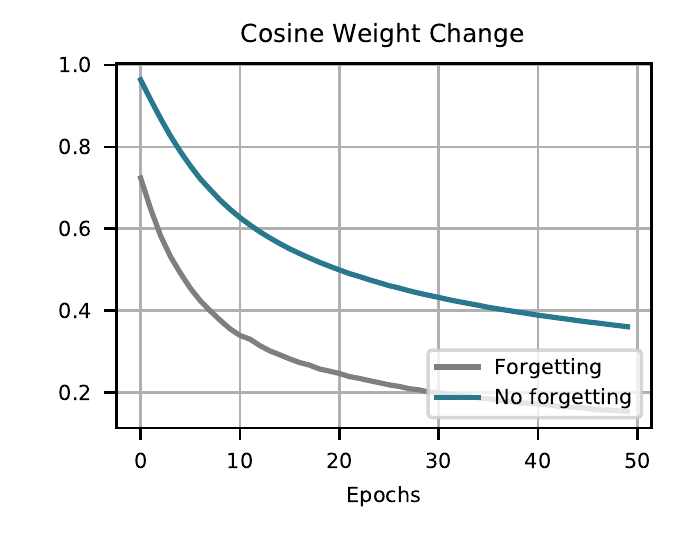}
         \label{fig:cosdist}
     \end{subfigure}
         \vspace*{-8mm}
    \caption{\textbf{Stabilizing final layer weights is not necessary to stop forgetting}. Here we look at the change in final layer weights during second task training for a model with severe forgetting (gray) and forgetting mitigated by rotating the second task representations (teal). We see that final layer weights become less and less similar in both cases. The setup is identical to. Figure~\ref{fig:dialingsup}}\label{fig:weightchange}
\end{figure}

\paragraph{Stopping forgetting without stabilizing weights} We saw in Section~\ref{sec-forgetting-hidden-reps} that both EWC and replay buffers stabilize deeper network representations. As mentioned above, this need not be the case a priori. As an illustrative example, we construct a setup where sequential training leads to no forgetting despite a significant change to final layer weights. We consider the single head model of Section~\ref{sec:analytic} where the Task 1 and Task 2 representations are completely orthogonal, $\Theta(x,x')=0$ for $x\in X^{(1)}_{\textrm{test}}$ and $x'\in X^{(2)}_{\textrm{train}}$. In this case the Task 1 predictions remain constant throughout all of Task 2 training, despite significant changes to the final layer weights and predictions on the second task. In Figure~\ref{fig:weightchange} we present an example of this setup.  The representations are again made orthogonal by explicitly rotating the second task features by the rotation matrix, $R(\pi/2)$ defined in Equation~\eqref{eq:rotationdef}.

\begin{figure}[h]
\centering
\includegraphics[width=12cm]{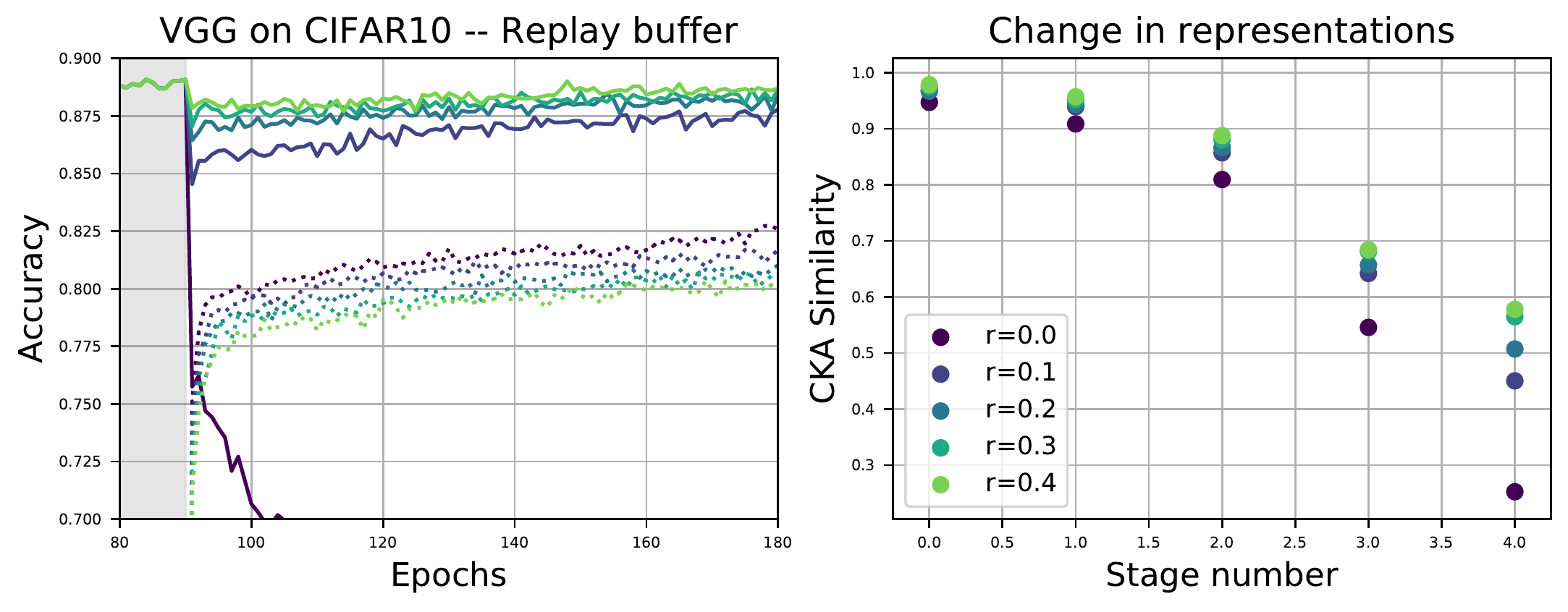}
\caption{\textbf{VGG Replay buffer CIFAR-10}.}
\end{figure}

\begin{figure}[h]
\centering
\includegraphics[width=12cm]{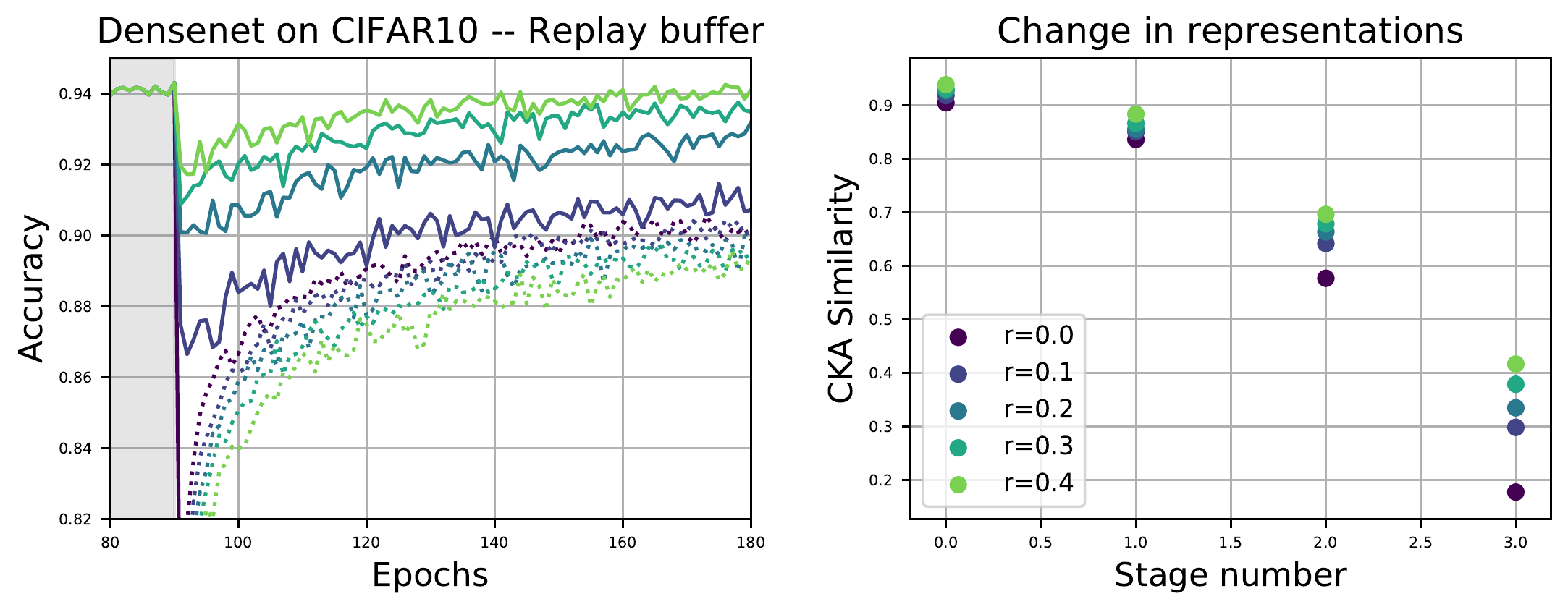}
\caption{\textbf{DenseNet Replay buffer CIFAR-10}.}
\end{figure}

\begin{figure}[h]
\centering
\includegraphics[width=12cm]{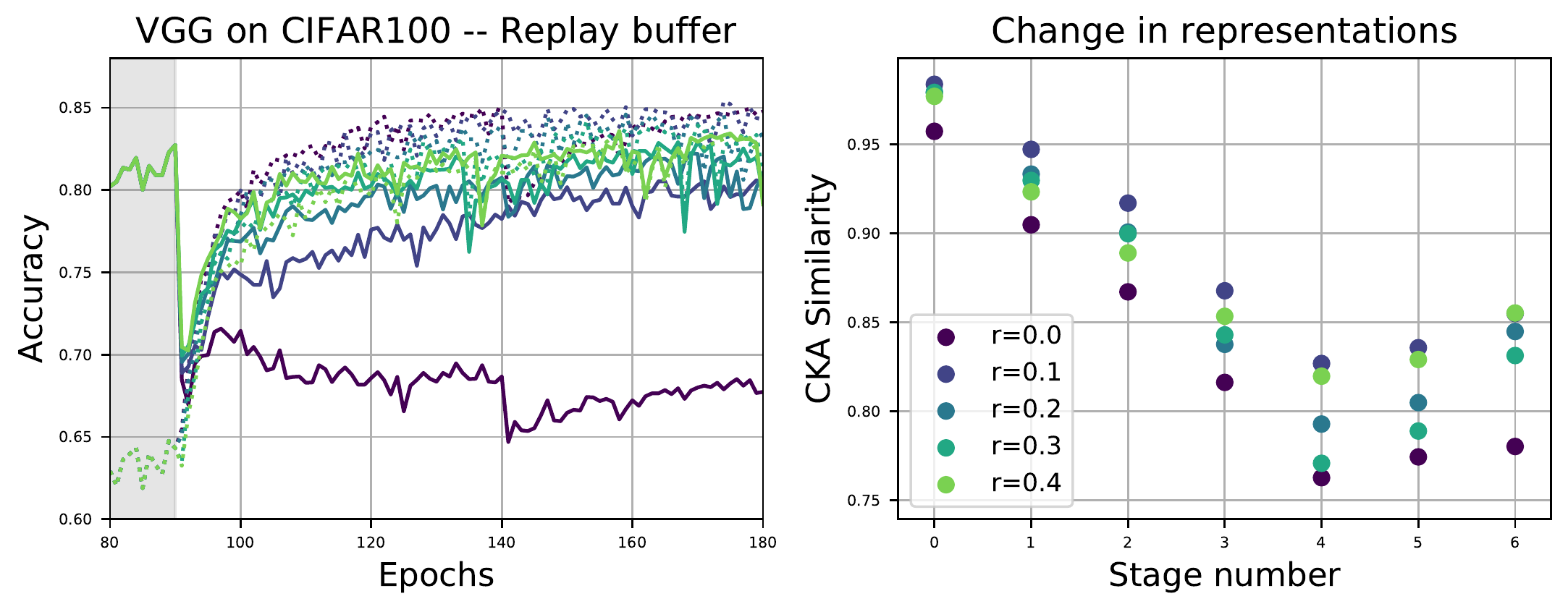}
\caption{\textbf{VGG Replay buffer CIFAR-100}.}
\end{figure}

\begin{figure}[h]
\centering
\includegraphics[width=12cm]{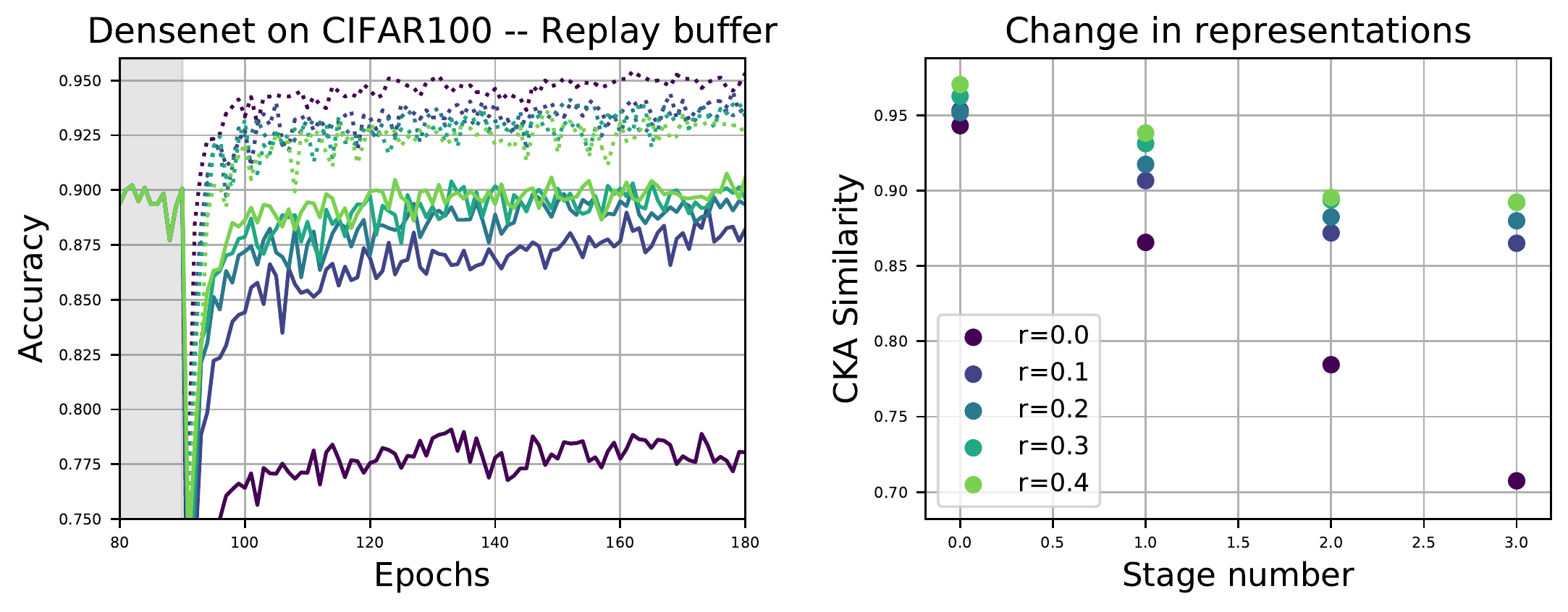}
\caption{\textbf{DenseNet Replay buffer CIFAR-100}.}
\end{figure}

\begin{figure}[h]
\centering
\includegraphics[width=12cm]{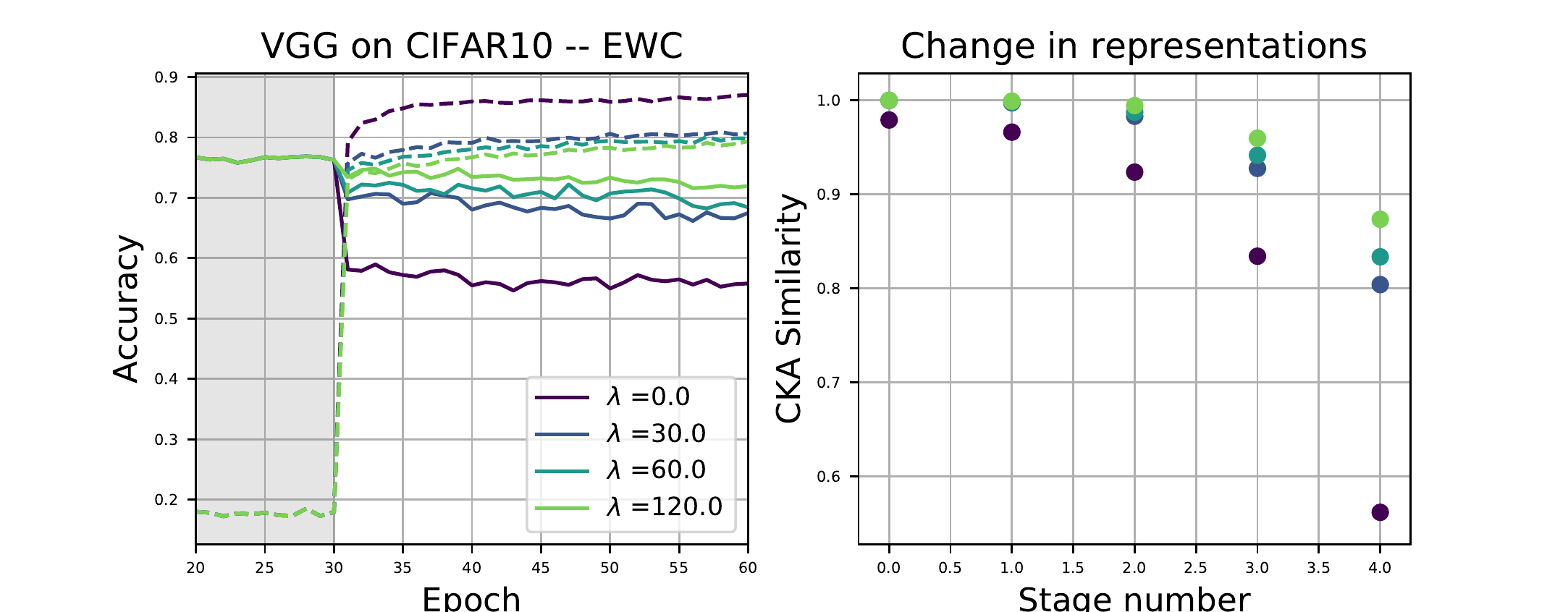}
\caption{\textbf{VGG EWC CIFAR-10}.}
\end{figure}

\begin{figure}[h]
\centering
\includegraphics[width=12cm]{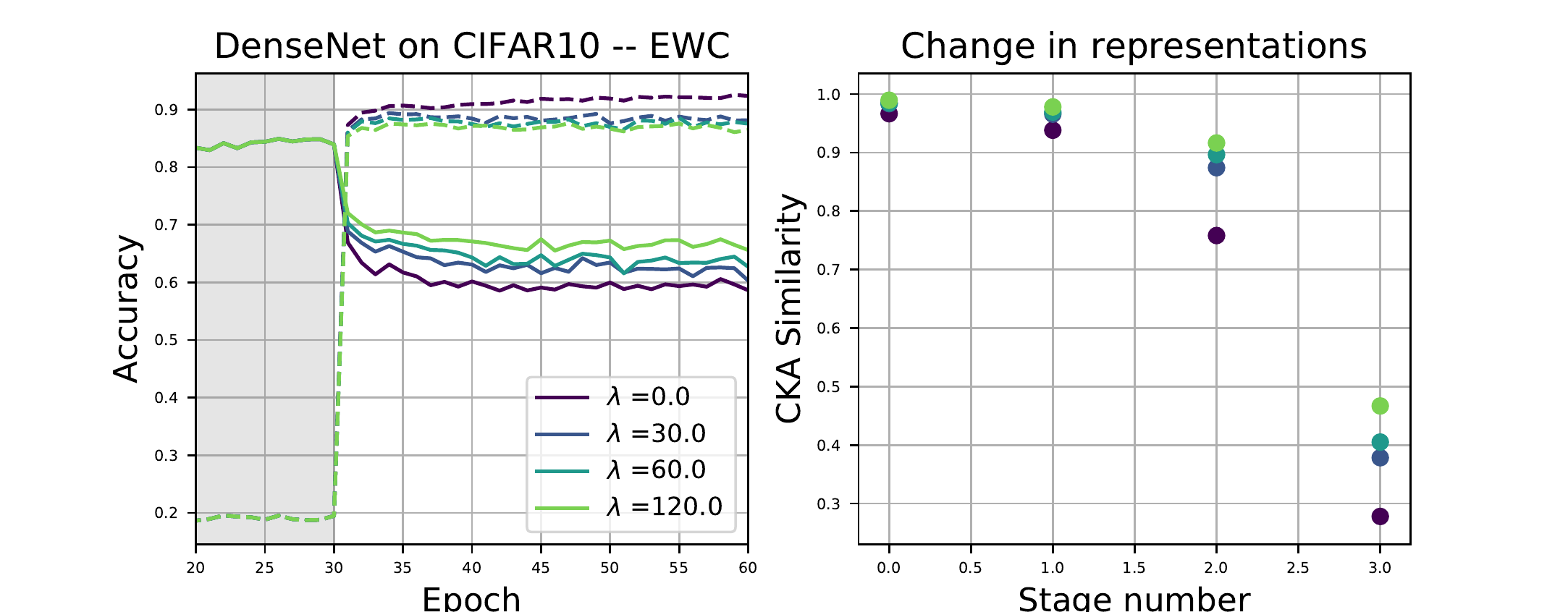}
\caption{\textbf{DenseNet EWC CIFAR-10}.}
\end{figure}

\begin{figure}[h]
\centering
\includegraphics[width=12cm]{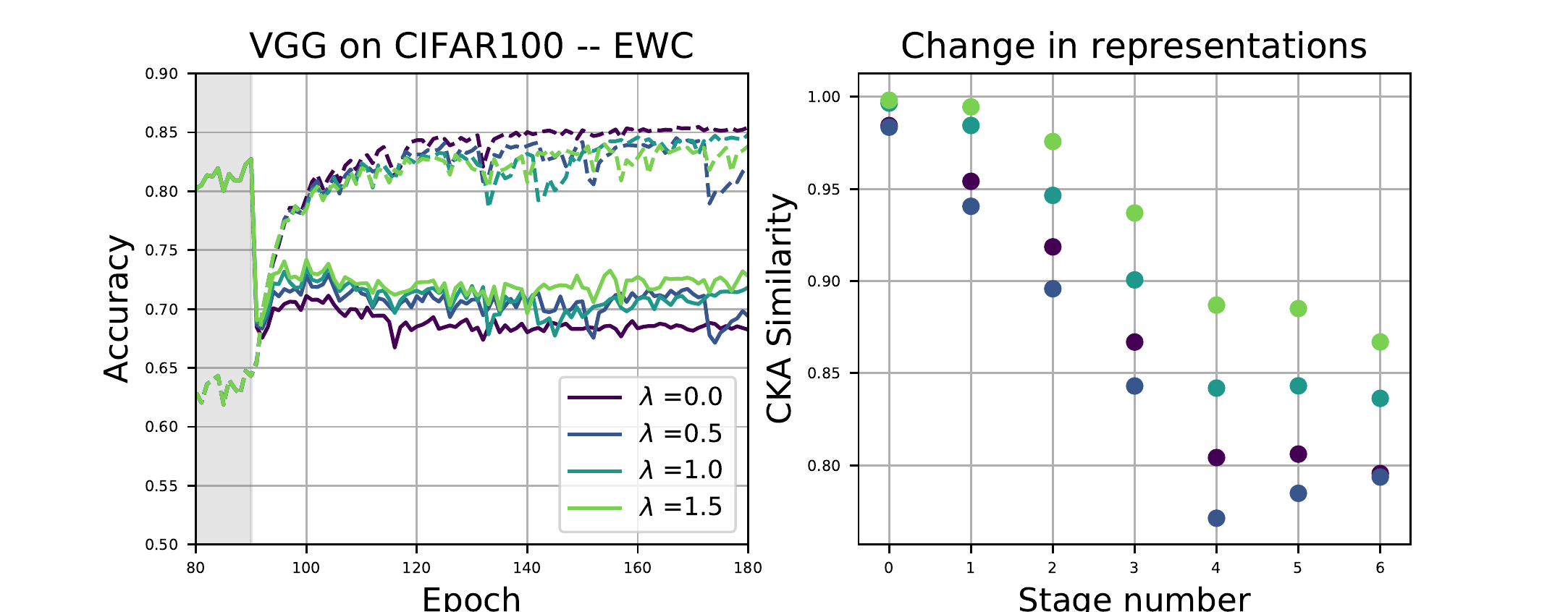}
\caption{\textbf{VGG EWC CIFAR-100}.}
\end{figure}

\begin{figure}[h]
\centering
\includegraphics[width=12cm]{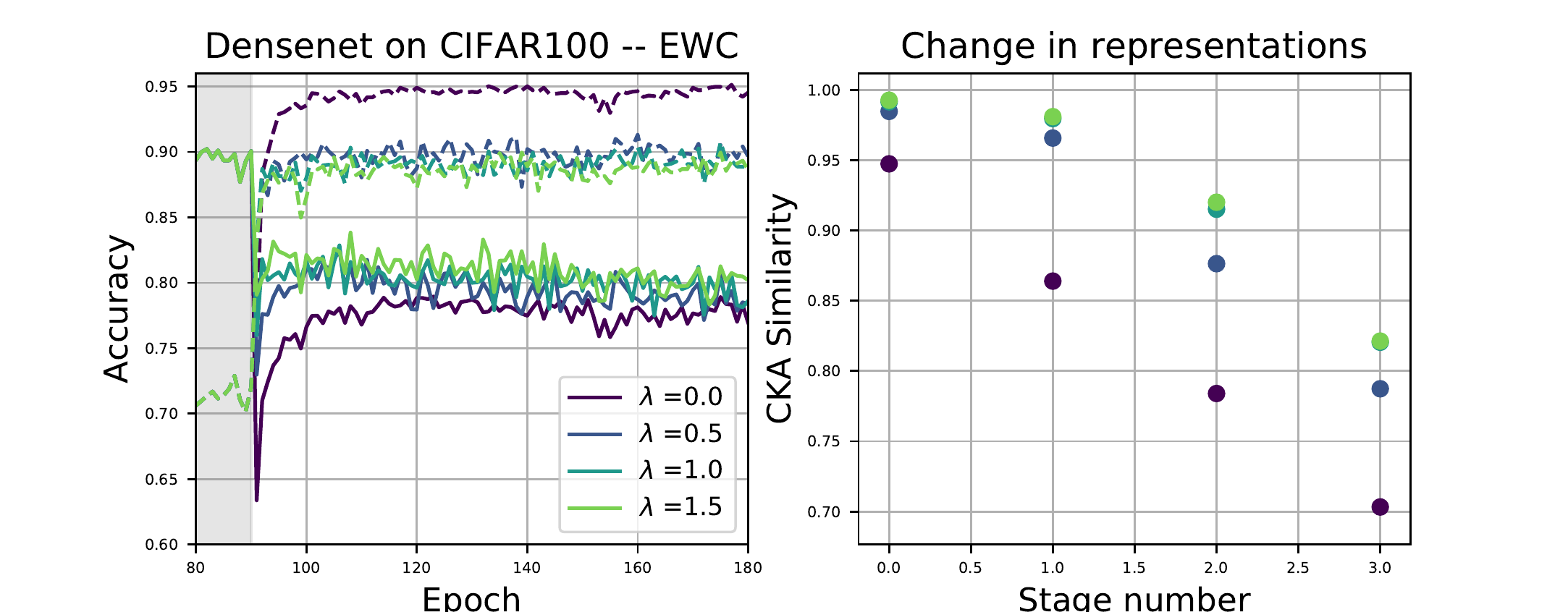}
\caption{\textbf{DenseNet EWC CIFAR-100}.}
\end{figure}


\end{document}